\documentclass[conference]{IEEEtran}

\usepackage[english]{babel}                         %
\usepackage{graphicx}                               %
\usepackage{hyperref}                               %
\usepackage[backend=biber, style=ieee]{biblatex}    %
\usepackage{csquotes}                               %
\usepackage{bm}                                     %
\usepackage[switch]{lineno}                         %
\usepackage[bibliography=common]{apxproof}          %
\usepackage{tabularx}
\usepackage{dirtytalk}
\usepackage{xparse}                                 %

\usepackage{notations}

\usepackage{algorithm} 

\usepackage[normalem]{ulem}

\newtheoremrep{theorem}{Theorem}
\newtheoremrep{lemma}{Lemma}
\newtheorem{definition}{Definition}
\newtheoremrep{corollary}{Corollary}

\graphicspath{{figures/}}

\setuptodonotes{disable}

\makeatletter
\renewcommand{\say}[1]%
  {%
    \addtocounter{dirtytalk@qdepth}{1}%
    \ifnum\thedirtytalk@qdepth=1\begin{quote}\fi%
    \dirtytalk@lsymb%
    \itshape
    #1%
    \dirtytalk@rsymb%
    \ifnum\thedirtytalk@qdepth=1\end{quote}\fi%
    \addtocounter{dirtytalk@qdepth}{-1}%
  }
\makeatother

\newcommand{\llabel}[1]{}
\newcommand{\lref}[1]{}

\title{Under manipulations, are some AI models\\ harder to audit?}

\makeatletter %
\newcommand{\linebreakand}{%
  \end{@IEEEauthorhalign}
  \hfill\mbox{}\par
  \mbox{}\hfill\begin{@IEEEauthorhalign}
}
\makeatother 

\IEEEoverridecommandlockouts
\author{%
    \IEEEauthorblockN{Augustin Godinot}
    \IEEEauthorblockA{\textit{Univ Rennes, Inria, CNRS, IRISA, PEReN} \\
    Rennes, France \\
    augustin.godinot@inria.fr}
    \and
    \IEEEauthorblockN{Erwan Le Merrer}
    \IEEEauthorblockA{\textit{Univ Rennes, Inria, CNRS, IRISA} \\
    Rennes, France}
    \and
    \IEEEauthorblockN{Gilles Tr\'edan}
    \IEEEauthorblockA{\textit{LAAS/CNRS} \\
    Toulouse, France}
    \and
    \linebreakand
    \IEEEauthorblockN{Camilla Penzo}
    \IEEEauthorblockA{\textit{PEReN} \\
    Paris, France}
    \and
    \IEEEauthorblockN{François Ta\"iani}
    \IEEEauthorblockA{\textit{Univ Rennes, Inria, CNRS, IRISA} \\
    Rennes, France}
}

\begin{document}

\maketitle

\begin{abstract}
    Auditors need robust methods to assess the compliance of web platforms with the
    law. However, since they hardly ever have access to the algorithm,
    implementation, or training data used by a platform, the problem is harder than
    a simple metric estimation. Within the recent framework of
    \emph{manipulation-proof} auditing, we study in this paper the feasibility of
    robust audits in realistic settings, in which models exhibit large capacities.

    We first prove a constraining result: if a web platform uses models that may fit
    any data, no audit strategy---whether active or not---can outperform random
    sampling when estimating properties such as demographic parity. To better
    understand the conditions under which state-of-the-art auditing techniques may
    remain competitive, we then relate the manipulability of audits to the capacity
    of the targeted models, using the Rademacher complexity. We empirically validate
    these results on popular models of increasing capacities, thus confirming
    experimentally that large-capacity models, which are commonly used in practice,
    are particularly hard to audit robustly. These results refine the limits of the
    auditing problem, and open up enticing questions on the connection between model
    capacity and the ability of platforms to manipulate audit attempts.
\end{abstract}


\vspace{2em}

\section{Introduction}

The pervasive deployment of user-facing automated decision systems raises
concerns over their impact on society. The growing number of online platforms
and their increasing complexity highlights the need for automated and robust
audits to assess their impact on users. The advent of highly publicized
audits---such as ProPublica's story on COMPAS \cite{larsonHowWeAnalyzed2016} or
Reuters' study on Amazon's recruiting tool
\cite{dastinAmazonScrapsSecret2018}---has brought considerable traction to the
AI audit field. For the public to trust Artificial Intelligence (AI) systems,
and more broadly algorithmic decision systems, we need methods to explain the
decisions of such systems
\cite{ribeiroWhyShouldTrust2016,lundbergUnifiedApproachInterpreting2017},
certify their implementation
\cite{yanActiveFairnessAuditing2022,shamsabadiConfidentialPROFITTConfidentialPROof2023}
and automatically and robustly detect misconduct
\cite{matiasSoftwareSupportedAuditsDecisionMaking2022,rastegarpanahAuditingBlackBoxPrediction2021}.

As it is common in the literature (e.g. see
~\cite{yanActiveFairnessAuditing2022}), we assume that the system is composed of
a \emph{trained machine learning (ML) model} $h$ that the auditor can interact
with via a web interface or an Application Programming Interface (API).
Similarly to the \emph{honest-but-curious (HBC)} threat model
\cite{goldreichFoundationsCryptographyVolume2009}, the outputs returned by the
API are the actual output of $h$. However, while the platform cannot arbitrarily
directly modify the output of $h$, it can use the interactions performed by the
auditor during the audit process to acquire as much information as possible and
modify the model in its favor. In this work, we focus on \emph{external
    certification audits}. In this type of audit, an external auditor (e.g. a
regulator or an auditing company) needs to verify a given property $\mu$ (e.g.
the absence of bias) of the API provided by the platform. We will refer to this
setting as the \emph{remote black-box auditing} problem.

\begin{figure}
    \centering
    \includegraphics[width=\columnwidth]{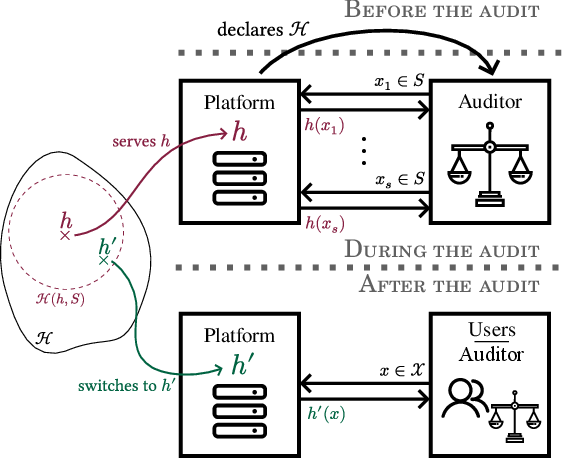}
    \caption{
        Security game of the manipulation-proof auditing framework. Before the audit,
        the platform declares the hypothesis space $\hypotspace$ to the auditor. During
        the audit, the platform serves the model $h \in \hypotspace$ and the auditor
        queries $h$ on $S$. After the audit, the platform can change its model to
        $h^\prime$ with the constraint that $\forall x \in S, h^\prime(x) = h(x)$ or
        equivalently, $h^\prime \in \hypotspace (h, S)$. } \label{fig:threat_model}
\end{figure}

Most of the current audit methods
\cite{metaxaAuditingAlgorithmsUnderstanding2021,sandvigAuditingAlgorithmsResearch2014}
could be referred to as \enquote{detection} audits. This is because they seek to
detect whether some rule is being violated either to improve the platform itself
or to take legal action. A typical methodology of \enquote{detection} audits
consists in randomly sampling the input space, computing the measure(s) of
interest and declaring the audit failed if the measures cross a given threshold.
In this case, to prove the platform's misconduct, one must witness it during the
audit. As a result, proving the absence of misconduct would require probing the
entire input space of the model $h$. Since the auditor cannot query the model on
its entire input space $\samplespace$, they must choose a subset $S\subseteq
    \samplespace$, and they must have the guarantee that the estimation on the
subset $S$ is not \enquote{too far} from the value they would find if they could
sample the whole input space.

\paragraph*{Threat model} \llabel{l:mp_intuitive_description} We describe the
interaction between the auditor and the platform in the threat model diagram
\autoref{fig:threat_model}. Before the audit, the platform discloses the
hypothesis space $\hypotspace$ they use (decision trees for example) to the
auditor. Then, during the auditing phase, the auditor interacts with the
(unknown) model $h \in \hypotspace$ exposed by the platform to iteratively build
an audit set $S \subset \samplespace$. The manipulation-proof framework
acknowledges the possibility for a platform to try to \emph{evade} the audit by
showing a fair model $h$ to the auditor, then switching to a more accurate but
potentially unfair model $h^\prime$. The only assumption on how the platform may
choose the new model $h^\prime$ is that it should be \emph{consistent} with $h$.
The consistency constraint requires $h^\prime$ to have the same outputs as $h$
on the audit set $S$, otherwise the auditor could easily check that the platform
changed its model after the audit by re-querying it on $S$. We now formalize the
capabilities and knowledge of the platform and the auditor in the manipulation
proof framework. \llabel{l:formal_threat_model}\begin{itemize}
    \item \emph{Auditor capabilities}: The auditor can send adaptive queries to the
          platform to build an audit set $S \subset \samplespace$.
    \item \emph{Auditor knowledge}: The auditor knows the hypothesis class $\hypotspace$
          implemented by the platform and the value of the sensitive attribute $x_A$ of
          all the points in the input space $\samplespace$. However, the auditor does not
          know the specific hypothesis $h \in \hypotspace$ implemented by the platform.
    \item \emph{Platform capabilities}: The platform can change its model from $h \in
              \hypotspace$ to $h^\prime \in \hypotspace$ after the audit as long as $h^\prime$
          respects the consistency constraint $\forall x \in S, h(x) = h^\prime(x)$.
    \item \emph{Platform knowledge}: The platform knows the property $\mu$ (e.g.
          Demographic Parity) being measured by the auditor. As the auditor, it knows the
          value of the sensitive attribute $x_A$ of all the points in the input space
          $\samplespace$.
\end{itemize}

\paragraph*{Problem} Among the attempts at formalizing robust auditing
\cite{yanActiveFairnessAuditing2022,yadavXAuditTheoreticalLook2023,chuggAuditingFairnessBetting2023a},
\citeauthor{yanActiveFairnessAuditing2022}~\cite{yanActiveFairnessAuditing2022}
have shown that the knowledge of the hypothesis class used by the platform can
\emph{potentially} reduce the required number of audit queries to reach a given
robustness level. Their method is based on disagreement-based active learning
\cite{hannekeTheoryDisagreementBasedActive2014} which requires training
surrogates of the platform's model. However, they only demonstrated their
proposed audit algorithm with linear models on small datasets
(\student~\cite{cortezUsingDataMining2008} and
\compas~\cite{larsonHowWeAnalyzed2016}). Furthermore, they prove that
quantifying the potential improvement (in terms of query complexity) of their
algorithm over a simple random baseline is computationally intractable. Thus,
whether it is possible or not to devise practical robust auditing methods still
remains an open question.

\llabel{l:scope}
Our exploration of robust audits for practical models is focused on binary
classifiers and binary sensitive attributes. While this calls for future work on other tasks and modalities,
this first exploration covers a large class of decision systems based on ML
algorithms \cite{richardsonDefiningDemystifyingAutomated2021}. \llabel{l:motivation} Our hope is to
demonstrate that regulators should be given more than black-box access to AI
models as part of the audit procedure.

\paragraph*{Contributions}

In this work, we investigate whether the platform can engineer models that
simultaneously achieve a high utility and evade the audit. To that end, we
compare the manipulation-proofness (MP) guarantees of a simple uniform random
audit algorithm (\autoref{algo:random_audit}) against the best guarantees a
regulator could hope for. Our contributions are three-fold.

\llabel{l:contributions}\begin{enumerate}
    \item We first consider those hypothesis classes that can perfectly reproduce any
          labeling of the dataset. This covers two practical cases: either the platform
          has a model with a very high capacity, or the auditor's prior on the platform's
          model is uninformative. We prove in
          \autoref{thm:shattering_implies_point_equivalence}
          (Subsection~\ref{subsec:infinite_capacity_models}) that no audit
          method---whether active or passive---can deliver a better performance than
          random sampling. We also prove in
          \autoref{corr:sensitive_attr_predictability_implies_point_equivalence} that this
          impossibility holds even if the hypothesis class can only imperfectly reproduce
          any labeling of the dataset with a bounded error rate.

    \item To uncover what properties of the hypothesis class influence its auditability,
          in Subsection~\ref{subsec:dictionaries} we analyze the simple class of
          dictionary models, whose manipulation guarantees can be analytically derived. We
          identify regimes in which the hypothesis class cannot be audited more
          efficiently than by random sampling.

    \item To build a practical understanding of our theoretical results, we formally
          define the notion of \emph{manipulability under random audits} and
          \emph{capacity} in \autoref{subsec:difficulty_capacity}. We then evaluate the
          manipulability under random audits of classical ML models for tabular data. We
          empirically confirm the strong connection between the classical \emph{Rademacher
              complexity} and the manipulability of manipulation-proof auditing. Since modern
          ML hypothesis classes tend to exhibit larger and larger capacities, we argue
          that our work brings up the limits of the current formulation of
          manipulation-proof auditing.
\end{enumerate}

\section{Auditing and manipulation-proof estimation}\label{sec:problem_setting}

During a typical audit, the auditor defines a measure of interest $\mu$ with an
associated threshold $\tau_\mu$. Classical measures used by auditors are
statistical parity indicators \cite{barocasFairnessMachineLearning2023} focusing
on independence (e.g. demographic parity, group fairness), separation (e.g.
balance for positive/negative class, equalized odds) and sufficiency (e.g.
calibration, predictive parity). Given that demographic parity does not require
any ground truth labels and since it is often used as the archetypal example in
the literature, we use it as the measure $\mu$ throughout this paper. While the
results we present refer specifically to demographic parity, it is
straightforward to extend them to any parity measure of the form%
\begin{align}
    \mu(h,S) = & \probap{h(X) = 1 \given X \in S,  E}                   \\
               & - \probap{h(X) = 1 \given X \in S, \snot{E}} \nonumber
\end{align}
with $E$ an event defined with respect to the random variables $X, X_A$ and $Y$,
where $X$ represents the input, $Y$ the ground truth label, and $X_A\in \{0,1\}$
is the sensitive attribute of interest for the auditor. For example, for
demographic parity, $E = (X_A = 1)$. We would like to stress that for other less
common measures that can nonetheless present an interest for auditors (e.g.
level of privacy \cite{luGeneralFrameworkAuditing2022} or the degree of
compliance with data minimization
\cite{rastegarpanahAuditingBlackBoxPrediction2021}), manipulation proof auditing
remains an open problem.

\subsection{Machine Learning notations}\label{subsec:ml_notations}

Except when noted, we will consider a binary classification task as in
\cite{dasguptaTeachingBlackboxLearner2019}, with finite \emph{input space}
$\samplespace$ and output space $\outputspace = \set{0, 1}$.\footnote{Should
    $\samplespace$ be infinite, \citeauthor{dasguptaTeachingBlackboxLearner2019}
    \cite{dasguptaTeachingBlackboxLearner2019} note that it suffices to sample a
    finite i.i.d. subset $\widetilde{\samplespace}$ and extend all the following
    bounds by classical generalization bounds.} $\outputspace^{\samplespace}$
denotes the space of functions $\samplespace \to \outputspace$. For any sample
$x \in \samplespace$, we refer to its sensitive attribute (e.g., gender,
ethnicity, religion) as $x_A\in\set{0, 1}$. The sensitive attribute of the
points in $\samplespace$ induces a partition of the input space. We note
$\samplespace_A = \set{x \in \samplespace : x_A = 1}$ and remark that
$\samplespace_{\snot{A}} = \snot{\samplespace_A}$. For any set $V$,
$\subsetsp{V}$ denotes the set of all subsets of $V$ and $\mathcal{U}(V)$
denotes the uniform distribution on $V$. By training the classification model,
the platform effectively chooses a model $h$ in some hypothesis class
$\hypotspace$. The auditor defines a measure $\mu : \hypotspace \times
    \subsetsp{\samplespace} \to \R_+$, which is known by the platform. For any
subset $V \subseteq \hypotspace$ and $S \subseteq \samplespace$, we define
\emph{the diameter of $V$ with respect to the measure $\mu$} as %
\begin{equation*}
    \mudiam[\mu(\cdot, S)] V = \max_{h, h^\prime \in V} |\mu(h, S) - \mu(h^\prime,S)|,
\end{equation*}%
when $S$ is the entire input space $\samplespace$, we abuse the notation and
write $\mudiam[\mu(\cdot, \samplespace)] V = \mudiam V$. Finally, define for any
subset $V \subset \hypotspace$, sample $x \in \samplespace$ and label $y \in
    \set{0,1}$ the set $V [x, y] = \set{h \in V : h(x) = y}$. The cost $\cost(V)$ of
a subset $V \subset \hypotspace$ is defined in \autoref{eq:auditing_cost}. Note
that when the context is clear, we elide the $\epsilon$ for simplicity. %
\begin{equation}\label{eq:auditing_cost}
    \text{Cost}_\epsilon(V) = \begin{cases}
        0  \text{ if } \mudiam V < \epsilon \\
        1 + \min\limits_{x \in \samplespace} \max\limits_{y \in \set{0,1}} \; \text{Cost}_\epsilon (V [x, y]) \text{ else}
    \end{cases}
\end{equation}

Before we formally define the \emph{capacity} of a hypothesis class in
\autoref{subsec:model_capacity}, we will use the term capacity loosely.
Intuitively the capacity of a hypothesis class $\hypotspace$ is related to the
ability for any labeling of the input space $\samplespace$ to find a hypothesis
$h \in \hypotspace$ that realizes this labeling. More details on the notion of
capacity can be found in \autoref{sec:related_work}.

\subsection{What is an active auditing algorithm?}\label{subsec:audit_algorithm}

An audit algorithm $\algo$ with label budget $s$ is a sequence of (possibly
randomized) $s + 1 $ functions $(f_0, \dots, f_s)$. For each iteration $i$, the
function $f_i : (\samplespace \times \set{0, 1})^{i+1} \to \samplespace$ chooses
the next sample $x_i = f_i \left(\left( x_0, h(x_0) \right), \dots, \left(
    x_{i-1}, h(x_{i-1}) \right) \right)$ to query and add to the audit set. After
the query budget has been spent, the end result of the algorithm is the audit
set $S=\algo(h)$. Note that most published black-box audits of web platforms are
not active \cite{bandyProblematicMachineBehavior2021}. In this case, an audit
algorithm reduces to a single (possibly randomized) function $f_s$ which does
not depend on the answers provided by the platform.

\subsection{The manipulation-proof auditing framework}
Following the framework of \citeauthor{yanActiveFairnessAuditing2022}
\cite{yanActiveFairnessAuditing2022}, the platform is assumed to be
\emph{self-consistent}, i.e. when the platform returns a given output $y = h(x)$
to an auditor's query $x$, the platform commits to this value and cannot return
a different answer $y^\prime = h(x)$ if $x$ is queried again at a later moment
in time. Furthermore, as explained in the threat model
\autoref{fig:threat_model}, it is assumed that the auditor knows the
\emph{hypothesis class} $\hypotspace \subseteq \set{0, 1}^\samplespace$ of the
model implemented by the platform. The self-consistency of the platform together
with the knowledge of the hypothesis class defines a subset of
\enquote{plausible} models in $\hypotspace$ that have the same answers as the
platform on the current audit set $S$. This subset is called the \emph{version
    space of $\hypotspace$ induced by $S$ and
    $h$}~\cite{mitchellGeneralizationSearch1982,hannekeTheoryDisagreementBasedActive2014},
noted $\hypotspace(h, S)$.%
\begin{equation}\label{eq:version_space}
    \hypotspace(h, S) = \set{h^\prime \in \hypotspace \sut \forall x \in S, h^\prime(x) = h(x)}
\end{equation}

We assume that the platform seeks to maximize its profits, which is not
necessarily aligned with the property that the regulator seeks to enforce.
During the audit process, the auditor incrementally builds an audit set $S
    \subseteq \samplespace$ based on their previous queries and the answers of the
platform. The goal of the auditor is to produce an estimate $\widehat{\mu}$ as
close as possible to the real value while being robust to the potential
manipulations implemented by the platform. We now formulate the two requirements
of the \emph{manipulation-proof} auditing problem, as introduced in
\cite{yanActiveFairnessAuditing2022}. \llabel{l:mp_audit_formulation}%
\begin{align}
    \text{Create an algorithm } \algo \text{ with} & \text{ smallest budget s}  \text{ such that,}     \nonumber                                \\
    \text{\textit{(fidelity)}\quad}                & |\mu(h, \algo(h)) - \mu(h, \samplespace)| < \epsilon  \label{eq:manipulationproof_audit_1} \\
    \text{\textit{(manipulation-proofness)}\quad}  & \mudiam \versionspacep{h, \algo(h)} <    \epsilon \label{eq:manipulationproof_audit_2}
\end{align}

\emph{Fidelity} is the classical estimation constraints. It requires the
estimated value $\widehat{\mu} = \mu(h^*, S)$ to be close to the real value
$\mu(h, \samplespace)$. In addition, \emph{manipulation-proofness} requires that
if the platform changes its implemented instance from $h$ to $h^\prime$ while
respecting the self-consistency constraint $h^\prime \in \versionspacep{h, S}$,
the difference between the previous $\mu(h, \samplespace)$ and new
$\mu(h^\prime, \samplespace)$ values of $\mu$ must be bounded. Therefore, the
$\mu$-diameter is the biggest change in the value of $\mu$ the auditor would
accept if the platform changed to another (consistent) hypothesis.

\subsection{Comparing manipulation-proof auditing algorithms}\label{subsec:comparing_audit_algorithms}

\begin{table}
    \centering
    \renewcommand{\arraystretch}{1.3}
    \caption{
        The query complexity of different auditing algorithms in the manipulation-proof
        framework, extracted from \citeauthor{yanActiveFairnessAuditing2022}
        \cite{yanActiveFairnessAuditing2022} } \label{tab:query_complexities}
    \begin{tabularx}{\columnwidth}{>{\raggedright\arraybackslash}X | p{3cm}}
        Algorithm                                                                   & Query complexity                  \\ \hline Random sampling
        (\autoref{algo:random_audit})                                               & $\bigo\left( \frac{1}{\epsilon^2}
        \log\left|\mathcal{H}\right|\right)$                                                                            \\ Optimal deterministic \cite[Algorithm
        1]{yanActiveFairnessAuditing2022}                                           & $\cost_\epsilon (\hypotspace)$    \\ Oracle
        based approximation (AFA) \cite[Algorithm 3]{yanActiveFairnessAuditing2022} &
        $\bigo\left(\log\card{\hypotspace} \log\card{\samplespace}
            \cost(\hypotspace)\right)$
    \end{tabularx}
\end{table}

There are two ways to compare two audit algorithms $\algo$ and $\algo^\prime$.
Either fix a target manipulation-proofness guarantee $\epsilon$ and evaluate the
number of queries needed by $\algo$ and $\algo^\prime$, or fix the audit budget
$s$ and evaluate the $\mu$-diameter of the audit sets built by $\algo$ and
$\algo^\prime$.

\Citeauthor{yanActiveFairnessAuditing2022}~\cite{yanActiveFairnessAuditing2022}
focused on the former: the study of the \emph{query complexity} of different
audit algorithms. For general hypothesis classes, they introduced three auditing
algorithms. The first one is the baseline random audit algorithm. This audit
algorithm consists in sampling among points with positive and negative sensitive
attributes, and computing the empirical frequencies of the events $(h(X) = 1 |
    X_A = 1)$ and $(h(X) = 1 | X_A = 0)$ (see Algorithm \ref{algo:random_audit}). To
capture the minimal query complexity attainable by deterministic audit
algorithms, they introduced a second algorithm based on the recursive
minimization of $\cost(\hypotspace)$. Finally,
\citeauthor{yanActiveFairnessAuditing2022} introduced a third, oracle-based,
algorithm that we coin AFA. We summarize the query complexities proved by
\cite{yanActiveFairnessAuditing2022} in \autoref{tab:query_complexities}.

Motivated by the implementation of MP audit algorithms, we choose to focus on
the second comparison approach: fixing an audit budget and evaluating the
$\mu$-diameter. This approach is better suited to our situation since in
practice, auditors have a limited query budget that would be agreed upon with
the platform prior to the audit.

\subsection{The computational complexity of manipulation-proof auditing}\label{subsec:computational_complexity}

As exposed in \autoref{tab:query_complexities}, the best attainable query
complexity, as well as the query complexity of the more practical AFA algorithm
depend on the value of $\cost(\hypotspace)$. In addition, the computational
complexity of AFA \cite[Algorithm 1]{yanActiveFairnessAuditing2022} is the time
to train a model from the hypothesis class $\hypotspace$ multiplied by the query
complexity. However, \citeauthor{yanActiveFairnessAuditing2022} prove that
$\text{Cost}(\mathcal{H})$ is hard to compute, hard to approximate and hard to
optimize \cite[Proposition 3.5]{yanActiveFairnessAuditing2022}. Thus, not only
it prevents practical implementations of the optimal deterministic algorithm, it
also prevents practical analysis of the query complexity and computational
complexity of AFA for large models that are costly to train.

\section{The competitive effectiveness of random audits}\label{sec:theoretical_result}

Current state-of-the-art models for tabular data (see
\autoref{fig:mu-diam_vs_capacity}) and image data (see e.g.
\cite{zhangUnderstandingDeepLearning2021}) are able to fit very large train sets
with close to perfect accuracy while retaining good generalization properties.
In our setting this would mean that these models can represent any binary
classification function $f: \samplespace \to \set{0, 1}$ of the input space. As
we saw in \autoref{subsec:computational_complexity}, the only tractable
algorithm (AFA~\cite{yanActiveFairnessAuditing2022}) that was proposed to solve
the manipulation-proof auditing task (\autoref{eq:manipulationproof_audit_1} and
\ref{eq:manipulationproof_audit_2}) is still too computationally intense to
audit large models because it requires to be able to train a lot of copies
efficiently. Moreover, while \cite{yanActiveFairnessAuditing2022} experimented
on small datasets with linear models, there exists no implementations or
practical experiments on larger models. Thus, the potential gains brought by AFA
are hard to predict. Yet, for AFA to be used in practice, it would be necessary
to balance the extra cost induced by auditing with AFA with the added guarantees
of AFA. Thus, a natural practical question arises. \textbf{Is the added
    manipulation-proofness guarantee worth paying the computational toll?}

To answer this question, instead of analyzing $\text{Cost}(\mathcal{H})$ (which
is hard to compute and derive) as \cite{yanActiveFairnessAuditing2022}, we
directly express the value of $\mudiam \hypotspace(h, S)$ for specific
hypothesis classes. Identifying hypothesis classes $\hypotspace$ wherein the
value of $\mudiam(h, S)$ remains constant across all audit sets $S$ allows us to
find scenarios in which enhancing manipulation-proofness guarantees beyond that
of a random baseline is impossible.

In this section, we consider three typical but insightful forms of hypothesis
space $\hypotspace$ to better understand this balance between computational cost
and added robustness. We prove in \autoref{subsec:infinite_capacity_models} that
for hypothesis classes shattering the whole input space, all the audit
algorithms have the same performance as random sampling. Next, to understand
what happens for classes that are only able to fit a part of the dataset, we
consider the illustrative class $\dict_m$ of dictionaries of size $m$. We derive
the exact value of their $\mu$-diameter in \autoref{subsec:dictionaries} and
show the link between the memory as an intuitive notion of the capacity and the
MP guarantees obtainable when auditing dictionary models. Last but not least,
building on the results of \autoref{subsec:infinite_capacity_models} and
\autoref{subsec:dictionaries}, we introduce a formal notion of the capacity of a
binary classification hypothesis class as the maximum number of samples a
platform can interpolate while still retaining good generalization performance.
Under this definition, we prove in \autoref{subsec:large_capacity_models} that
large capacity models cannot be audited more efficiently than by the random
baseline.

\begin{algorithm}
    \begin{algorithmic}[1]
        \Require Proportions $\beta_1, \beta_2$, budget $s$
        \Ensure audit dataset $S$ with $\card{S} = s$
        \State $s^+ \gets \floorp{\beta_1 \card{\samplespace_A}}$, $s^- \gets \floorp{\beta_2 \card{\snot{\samplespace_A}}}$
        \State $S^+ \gets$ sample $s^+$ points in $\samplespace_A$ without replacement
        \State $S^- \gets$ sample $s^-$ points in $\snot{\samplespace_A}$ without replacement
        \State $S = S^+ \sqcup S^-$
    \end{algorithmic}
    \caption{A random sampling audit strategy}
    \label{algo:random_audit}
\end{algorithm}

\subsection{Hypothesis classes that can fit the dataset entirely}\label{subsec:infinite_capacity_models}

To build intuition on the following theorems, let us first consider classes able
to fit any distribution on $\samplespace$. This corresponds to the case of a
platform with a very large, overparametrized hypothesis class $\hypotspace$ able
to fit any labeling of the whole input space $\samplespace$. \footnote{ This
    does not contradict the No Free Lunch theorem since here, the input space
    $\samplespace$ is finite.} This assumption is equivalent to considering the
hypothesis class $\hypotspace = \set{0,1}^\samplespace$. Because all the
functions from the input space to the output space are possible, the answer of
the platform on a query $x$ does not give any information on the possible
answers to the other queries in $\samplespace$. It follows, that no matter how
the points are iteratively chosen, only the number of points (and the value of
their associated sensitive attribute) will matter in the computation of the
$\mu$-diameter. We now formalize this intuition.

\begin{theoremrep}[No need to aim] \label{thm:shattering_implies_point_equivalence}
    Let $\hypotspace = \set{0,1} ^ \samplespace$. For any audit set $S \subseteq
        \samplespace$ and hypothesis $h \in \hypotspace$,
    \begin{align*}
        \mudiam \hypotspace(h, S) = 2 - \big( & \probap{X \in S \given X_A = 1}        \\
                                              & + \probap{X \in S \given X_A = 0}\big)
    \end{align*}
\end{theoremrep}

\begin{proofsketch}
    The first step in proving \autoref{thm:shattering_implies_point_equivalence}
    relies on the fact that all the instances $h^\prime \in \versionspacep{h, S}$
    have the same value of $\mu(h^\prime, S)$. After decomposing the $\mu$-diameter
    on $S$ and $\snot{S}$, we use this fact to separate the $\mu$-diameter into the
    difference between a maximization and a minimization problem. The optima of
    these problems rely on the existence of hypotheses $h^\uparrow, h^\downarrow \in
        \versionspacep{h, S}$ that exactly fit the sensitive attribute (resp. its
    negation) on $\snot{S}$. Since $\hypotspace$ is the space of all functions, it
    is always possible to find such $h^\uparrow$ and $h^\downarrow$. Finally, we
    find these optima and simplify their expressions to reach that of
    \autoref{thm:shattering_implies_point_equivalence}. A complete proof is provided
    in \autoref{thm:shattering_implies_point_equivalence-apx}.
\end{proofsketch}

\begin{proof}
    The proof is executed in 4 steps: decomposition of the value of $\mu(h,
        \samplespace)$ on $S$ and $\snot{S}$, decomposition of the $\mu$-diameter on $S$
    and $\snot{S}$, solving the optimization on the decomposed problems and
    conclusion.

    \noindent\textbf{Step 1: Decompose $\mu$} \\
    For any $h \in \hypotspace$, $S \subseteq \samplespace$
    \begin{align}
        \mu(h, \samplespace) =\; & \probap{h(X) = 1 \given X_A = 1} - \probap{h(X) = 1 \given X_A = 0}                                                                                           \nonumber   \\
        = \;                     & \probap{h(X) = 1 \given X_A = 1, X \in S} \underbrace{\probap{X \in S \given X_A = 1}}_\alpha                                                                 \nonumber   \\
                                 & \quad + \probap{h(X) = 1 \given X_A = 1, X \in \snot{S}} \underbrace{\probap{X \in \snot{S} \given X_A = 1}}_{1 - \alpha}                                       \nonumber \\
                                 & \quad - \probap{h(X) = 1 \given X_A = 0, X \in S} \underbrace{\probap{X \in S \given X_A = 0}}_{\alpha - \delta}                                                \nonumber \\
                                 & \quad - \probap{h(X) = 1 \given X_A = 0, X \in \snot{S}} \underbrace{\probap{X \in \snot{S} \given X_A = 0}}_{1 - \alpha + \delta}                              \nonumber \\
        = \;                     & \alpha \underbrace{\big(\probap{h(X) = 1 \given X_A = 1, X \in S} - \probap{h(X) = 1 \given X_A = 0, X \in S}\big)}_{\mu(h, S)}                             \nonumber     \\
                                 & \quad + (1 - \alpha) \underbrace{\left(\probap{h(X) = 1 \given X_A = 1, X \in \snot{S}} - \probap{h(X) = 1 \given X_A = 0, X \in \snot{S}}\right)}_{\mu(h, S)}  \nonumber \\
                                 & \quad + \delta \left(\probap{h(X) = 1 \given X_A = 0, X \in S} - \probap{h(X) = 1 \given X_A = 0, X \in \snot{S}}\right)                                        \nonumber \\
        = \;                     & \alpha \mu(h, S) + (1 - \alpha) \mu(h, \snot{S})                                                                                                               \nonumber  \\
                                 & \quad + \delta \left(\probap{h(X) = 1 \given X_A = 0, X \in S} - \probap{h(X) = 1 \given X_A = 0, X \in \snot{S}}\right)
    \end{align}

    \noindent\textbf{Step 2: Decompose the $\mu$-diameter} \\
    For any $h, h^\prime \in \hypotspace(h^*, S)$,
    \begin{align}
        \mu(h, \samplespace) - \mu(h^\prime, \samplespace) =\; & \alpha \underbrace{\big(\mu(h, S) - \mu(h^\prime, S)\big)}_{= 0 \text{ since } h(S) = h^\prime(S) = h^*(S)} + (1 - \alpha) \left(\mu(h, \snot{S}) - \mu(h^\prime, \snot{S})\right)      \nonumber    \\
                                                               & \quad + \delta \underbrace{\big(\probap{h(X) = 1 \given X_A = 0, X \in S} - \probap{h^\prime(X) = 1 \given X_A = 0, X \in S}\big)}_{= 0 \text{ since } h(S) = h^\prime(S) = h^*(S)}     \nonumber    \\
                                                               & \quad + \delta \left(\probap{h^\prime(X) = 1 \given X_A = 0, X \in \snot{S}} - \probap{h(X) = 1 \given X_A = 0, X \in \snot{S}}\right)                                                     \nonumber
    \end{align}

    Using the definition and separability of the $\mu$-diameter, we have
    \begin{equation}
        \mudiamp{h^*, S} = \max_{h \in \hypotspace(h^*, S)} \mu(h, S) - \min_{h^\prime \in \hypotspace(h^*, S)} \mu(h, S)
    \end{equation}

    Therefore, by grouping the terms that depend on $h$ and $h^\prime$ in the
    previous development:
    \begin{align}
        \mudiamp{h^*, S} =\; & \colorbox{orange!20!white}{$\max_{h \in \hypotspace(h^*, S)} \Big[(1 - \alpha) \mu(h, \snot{S}) - \delta \probap{h(X) = 1 \given X_A = 0, X \in \snot{S}}\Big]$} \nonumber                                                \\
                             & - \colorbox{blue!20!white}{$\min_{h^\prime \in \hypotspace(h^*, S)} \Big[ (1 - \alpha) \mu(h^\prime, \snot{S}) - \delta \probap{h^\prime(X) = 1 \given X_A = 0, X \in \snot{S}}\Big]$}\label{eq:diam:as:max:min:on:not:S}
    \end{align}

    \noindent\textbf{Step 3: Solve each optimization problem} \\

    To solve the two optimization problems, we come back to the definition of $\mu$.
    \begin{align}
        \colorbox{orange!20!white}{\phantom{blind}} & = \max_{h \in \hypotspace(h^*, S)} \left\{(1 - \alpha) \probap{h(X) = 1 \given X_A = 1, X \in \snot{S}} - (1 - \alpha + \delta) \probap{h(X) = 1 \given X_A = 0, X \in \snot{S}}\right\}                          \\
                                                    & = - (1 - \alpha + \delta) + \max_{h \in \hypotspace(h^*, S)} \left\{(1 - \alpha) \probap{h(X) = 1 \given X_A = 1, X \in \snot{S}}+ (1 - \alpha + \delta) \probap{h(X) = 0 \given X_A = 0, X \in \snot{S}}\right\}
    \end{align}

    Similarly,
    \begin{align}
        \colorbox{blue!20!white}{\phantom{blind}} & = \min_{h \in \hypotspace(h^*, S)} \left\{(1 - \alpha) \probap{h(X) = 1 \given X_A = 1, X \in \snot{S}} - (1 - \alpha + \delta) \probap{h(X) = 1 \given X_A = 0, X \in \snot{S}}\right\}                          \\
                                                  & = - (1 - \alpha + \delta) + \min_{h \in \hypotspace(h^*, S)} \left\{(1 - \alpha) \probap{h(X) = 1 \given X_A = 1, X \in \snot{S}}+ (1 - \alpha + \delta) \probap{h(X) = 0 \given X_A = 0, X \in \snot{S}}\right\}
    \end{align}

    We write $h^\uparrow$ (resp. $h^\downarrow$) the minimizer of
    $\colorbox{orange!20!white}{\phantom{blind}}$ (resp.
    $\colorbox{blue!20!white}{\phantom{blind}}$).

    \begin{minipage}{.46\textwidth}
        \begin{equation}
            h^\uparrow(x) =
            \begin{cases}
                1 \text{  if } x_A = 1 \text{ and } x \in \snot{S} \\
                0 \text{  if } x_A = 0 \text{ and } x \in \snot{S} \\
                0 \text{  else}
            \end{cases}
        \end{equation}
    \end{minipage}%
    \hfill%
    \begin{minipage}{.46\textwidth}
        \begin{equation}
            h^\downarrow(x) =
            \begin{cases}
                1 \text{  if } x_A = 0 \text{ and } x \in \snot{S} \\
                0 \text{  if } x_A = 1 \text{ and } x \in \snot{S} \\
                0 \text{  else}
            \end{cases}
        \end{equation}
    \end{minipage}

    The optimizers $h^\uparrow$ and $h^\downarrow$ yield the optima
    \begin{align}
        \colorbox{orange!20!white}{\phantom{blind}} & = - (1 - \alpha + \delta) + (1 - \alpha) \underbrace{\probap{h^\uparrow(X) = 1 \given X_A = 1, X \in \snot{S}}}_{= 1} + (1 - \alpha + \delta) \underbrace{\probap{h^\uparrow(X) = 0 \given X_A = 0, X \in \snot{S}}}_{=1}      \\
                                                    & = 1 - \alpha                                                                                                                                                                                                                   \\
        \colorbox{blue!20!white}{\phantom{blind}}   & = - (1 - \alpha + \delta) + (1 - \alpha) \underbrace{\probap{h^\downarrow(X) = 1 \given X_A = 1, X \in \snot{S}}}_{=0} + (1 - \alpha + \delta) \underbrace{\probap{h^\downarrow(X) = 0 \given X_A = 0, X \in \snot{S}} }_{= 0} \\
                                                    & = -(1 - \alpha + \delta)
    \end{align}

    \noindent\textbf{Step 4: Conclusion} \\

    \begin{align}
        \mudiam \hypotspace(h^*, S) & = \colorbox{orange!20!white}{\phantom{blind}} - \colorbox{blue!20!white}{\phantom{blind}} \\
                                    & = (1 - \alpha) + (1 - \alpha + \delta)                                                    \\
                                    & =2 - \big(\probap{X \in S \given X_A = 1} + \probap{X \in S \given X_A = 0}\big)
    \end{align}
\end{proof}

The values $\probap{X \in S \given X_A = 1}$ and $\probap{X \in S \given X_A =
        0}$ are aggregated quantities that depend only on the relative proportion of
sensitive ($x_A = 1$) and non-sensitive ($x_A = 0$) samples in the audit set
$S$. Therefore, for any pair $(\probap{X \in S \given X_A = 1}, \probap{X \in S
        \given X_A =0})$, one can design a random sampling scheme that achieves the
desired relative proportions. We expose such algorithm in
\autoref{algo:random_audit}. Since the auditor by definition knows the sensitive
attribute of each sample, the idea is to sample points from $\samplespace_A$ and
$\samplespace_{\snot{A}}$ with the right proportions $(\beta_1, \beta_2)$ in
$S_\text{random}$. Setting $(\beta_1, \beta_2) = (\probap{X \in S \given X_A =
        1}, \probap{X \in S \given X_A =0})$ in \autoref{algo:random_audit} yields
$(\probap{X \in S_\text{random} \given X_A = 1}, \probap{X \in S_\text{random}
        \given X_A =0}) = (\probap{X \in S \given X_A = 1}, \probap{X \in S \given X_A
        =0})$. Following \autoref{thm:shattering_implies_point_equivalence}, any audit
set $S$ with the same relative proportions $(\probap{X \in S \given X_A = 1},
    \probap{X \in S \given X_A =0})$ yields the same $\mu$-diameter. Since any
couple $(\probap{X \in S \given X_A = 1}, \probap{X \in S \given X_A =0})$ is
also attainable by the random sampling algorithm described in
\autoref{algo:random_audit}, \textbf{when the hypothesis class can perfectly fit
    any arbitrary label distribution, all audit algorithms --active or not-- have at
    most the same manipulation-proofness guarantees as random sampling}.

As a side note, removing the assumption that the auditor knows the hypothesis
class implemented by the platform is equivalent to assuming $\hypotspace =
    \set{0, 1}^\samplespace$. In this sense, by proving that random sampling is
optimal when the hypothesis class is unknown,
\autoref{thm:shattering_implies_point_equivalence} demonstrates that knowing
$\hypotspace$ is necessary (but not sufficent) to design more efficient
manipulation-proof auditing methods.

\subsection{An illustrative example with dictionaries}\label{subsec:dictionaries}

\begin{figure}
    \centering
    \includegraphics{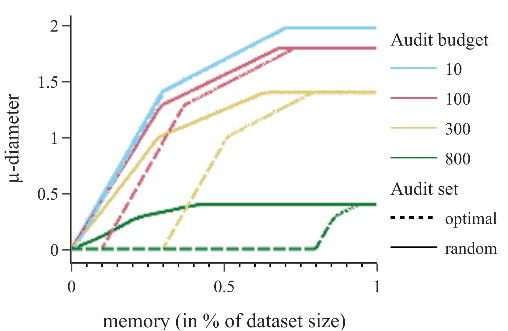}
    \caption{%
        The diameter (vertical axis) resulting from the amount of memory (horizontal
        axis) of the dictionary model studied in subsection \ref{subsec:dictionaries}.
        The various audit budgets are represented by different curve colors, while the
        optimal audit set appears as dashed curves, and the random baseline audit sets
        as plain lines. } \label{fig:diam-mem} %
\end{figure}

It is unlikely in practice that any hypothesis class can fit the entire input
space $\samplespace$. We now relax this assumption to pursue our analysis of the
achievable manipulation-proofness guarantees of models with a large capacity. To
that end, we introduce the class $\dict_m$ of dictionary models. A dictionary $d
    \in \dict_m$ is built by choosing a set of $m \in \intrvl{n}$ samples in
$\samplespace$ and storing the corresponding labels. When the dictionary is
asked to label a sample that it did not store, it returns $0$ as a default
value. Define, for any set of vectors $V \subseteq \R^d$, $\permset(V)$ the set
of vectors obtained from $V$ by including all permutations of the coefficients
of each $v \in V$. The hypothesis class of dictionaries of memory $m$ is
formally introduced in \autoref{def:dictionary}.

\begin{definition}[Dictionary hypothesis class]\label{def:dictionary}
    Consider an input space $\samplespace$, $n = \card{\samplespace}$. The class of
    dictionaries of memory $m \in \intrvl{n}$ is defined as%
    \begin{equation*}
        \dict_m = \permset\left( \set{0, 1}^m \times \set{0_{\R^{n-m}}}\right)
    \end{equation*}
\end{definition}

While such a hypothesis class is not likely to be used in a practical context
(as it will typically fail to generalize beyond the encountered examples,
exhibiting a blatant overfitting) it is simple enough to support an analysis of
the MP guarantees for both randomized and optimal approaches. Moreover, its main
parameter (the memory $m$) directly influences its capacity.

The exact value of the $\mu$-diameter of dictionary hypothesis classes is
exposed in \autoref{thm:mudiam_dicts}. The proof can be found in
\autoref{thm:mudiam_dicts-apx}.

\begin{theoremrep}[$\mu$-diameter of $\dict_m$]\label{thm:mudiam_dicts}
    Consider $S \subseteq \samplespace$, $d \in \dict_m$. Note $m^\prime = m -
        \card{x \in S : d(x) = 1}$. The $\mu$-diameter of $\dict_m(d, S)$ is given by
    \begin{equation*}
        \mudiam \dict_m(d, S) =
        \frac{
            \min(\card{\samplespace_A \cap \snot{S}}, m^\prime)
        }{\card{\samplespace_A}}
        + \frac{
            \min(\card{\snot{\samplespace_A} \cap \snot{S}}, m^\prime)
        }{\card{\snot{\samplespace_A}}}
    \end{equation*}
\end{theoremrep}

\begin{proofsketch}
    The proof relies on the same development of the diameter as in the proof of
    \autoref{thm:shattering_implies_point_equivalence} but instead of finding
    $h^\uparrow$ and $h^\downarrow$, we are able to give the values of the optima
    thanks to the structure of $\dict_m$.
\end{proofsketch}

\begin{proof}
    In the proof of \autoref{thm:shattering_implies_point_equivalence}, we
    established the following identity (for any hypothesis class thus for $\dict_m$,
    and for any $S$ and $d^*$):

    \begin{align}
         & \mudiam \dict(d^*, S)                                                                                                                                                                                                                                                \\
         & = \colorbox{orange!20!white}{$\max_{d \in \dict(d^*, S)} \left\{\probap{X \in \snot{S} \given X_A = 1} \probap{d(X) = 1 \given X_A = 1, X \in \snot{S}}+ \probap{X \in \snot{S} \given X_A = 0} \probap{d(X) = 0 \given X_A = 0, X \in \snot{S}}\right\}$} \nonumber \\
         & - \colorbox{blue!20!white}{$\min_{d \in \dict(d^*, S)} \left\{\probap{X \in \snot{S} \given X_A = 1} \probap{d(X) = 1 \given X_A = 1, X \in \snot{S}}+ \probap{X \in \snot{S} \given X_A = 0} \probap{d(X) = 0 \given X_A = 0, X \in \snot{S}}\right\}$}
    \end{align}

    First, observe that in the two optimization problems, the value of the objective
    function does not depend on the values of $d$ on $S$. Moreover, the choices of
    the labels $d(x)$ for $x \in \snot{S}$ can be made freely as long as $d$ does
    not have more than $m^\prime = m - \card{x \in S : d^*(x) = 1}$ \enquote{$1$}s
    (because it has to use $\card{x \in S : d^*(x) = 1}$ slots of memory to store
    the answers of $d^*$ on $S$).

    Therefore, the dictionary that optimizes
    \colorbox{orange!20!white}{\phantom{blind}} is built by storing as many
    \enquote{$1$}s in $d$ on the entries of $x \in \samplespace_A \cap \snot{S}$
    within the limits of the $m^\prime$ slots left. This leads to
    \begin{align}
        \colorbox{orange!20!white}{\phantom{blind}} =
        \probap{X \in \snot{S} \given X_A = 1} \frac{\min(\card{\samplespace_A \cap \snot{S}}, m^\prime)}{\card{\samplespace_A \cap \snot{S}}}
        + \probap{X \in \snot{S} \given X_A = 0} * 1
    \end{align}

    Next, rewriting as a maximization problem, we get
    \begin{align}
         & \colorbox{blue!20!white}{\phantom{blind}} =\probap{X \in \snot{S} \given X_A = 1} + \probap{X \in \snot{S} \given X_A = 0}                                                                                                    \nonumber \\
         & - \min_{d \in \dict(d^*, S)} \left\{\probap{X \in \snot{S} \given X_A = 1} \probap{d(X) = 0 \given X_A = 1, X \in \snot{S}} + \probap{X \in \snot{S} \given X_A = 0} \probap{d(X) = 1 \given X_A = 0, X \in \snot{S}}\right\}
    \end{align}

    Similar to the case of \colorbox{orange!20!white}{\phantom{blind}}, the
    dictionary that optimizes \colorbox{blue!20!white}{\phantom{blind}} is built by
    storing as many \enquote{$1$}s in $d$ on the entries of $x \in
        \snot{\samplespace_A} \cap \snot{S}$ withing the limits of the $m^\prime$ slots
    left. This leads to
    \begin{align}
        \colorbox{blue!20!white}{\phantom{blind}} & = \probap{X \in \snot{S} \given X_A = 1} + \probap{X \in \snot{S} \given X_A = 0}                                                                                                                          \\
                                                  & \qquad - \probap{X \in \snot{S} \given X_A = 1} * 1 - \probap{X \in \snot{S} \given X_A = 0} \frac{\min(\card{\snot{\samplespace_A} \cap \snot{S}}, m^\prime)}{\card{\snot{\samplespace_A} \cap \snot{S}}}
    \end{align}

    Composing the expressions of \colorbox{orange!20!white}{\phantom{blind}} and
    \colorbox{blue!20!white}{\phantom{blind}}, we get
    \begin{equation}
        \mudiam \dict(d^*, S) = \probap{X \in \snot{S} \given X_A = 1} \frac{\min(\card{\samplespace_A \cap \snot{S}}, m^\prime)}{\card{\samplespace_A \cap \snot{S}}} + \probap{X \in \snot{S} \given X_A = 0} \frac{\min(\card{\snot{\samplespace_A} \cap \snot{S}}, m^\prime)}{\card{\snot{\samplespace_A} \cap \snot{S}}}
    \end{equation}

    Here, it is important to understand that in the notations $\probap{X \in S}$ or
    $\probap{d(X) = 1}$, $X$ is a random variable taking values in $\samplespace$
    with a uniform probability. Therefore, $\probap{X \in \snot{S} \given X_A = 1} =
        \frac{\card{\samplespace_A \cap \snot{S}}}{\card{\samplespace_A}}$ and
    $\probap{X \in \snot{S} \given X_A = 0} = \frac{\card{\snot{\samplespace_A} \cap
                \snot{S}}}{\card{\snot{\samplespace_A}}}$, which simplifies the previous
    equation

    \begin{equation}
        \mudiam \dict(d^*, S) = \frac{\min(\card{\samplespace_A \cap \snot{S}}, m^\prime)}{\card{\samplespace_A}} + \frac{\min(\card{\snot{\samplespace_A} \cap \snot{S}}, m^\prime)}{\card{\snot{\samplespace_A}}}
    \end{equation}
\end{proof}

We are interested in the high memory $m$, low audit budget $\card{S}$ regime. In
this situation, there exist couples $(S, m)$ such that $\card{\samplespace_A
        \cap \snot{S}} \leq m^\prime$ and $\card{\snot{\samplespace_A} \cap \snot{S}}
    \leq m^\prime$. Thus, in this regime, the $\mu$-diameter does not depend on the
labels of the particular dictionary $d$ chosen by the platform. Therefore, as
for the case $\hypotspace = \set{0, 1}^\samplespace$, in the high memory, low
audit budget regime, all audit algorithms -- active or not -- have at most the
same manipulation-proofness guarantees as random sampling.

\paragraph*{Simulation of the impact of memory over diameter} The expression of the
$\mu$-diameter exposed in \autoref{thm:mudiam_dicts} is piecewise linear in the
memory $m$. To gain intuition, we plot the value of $\mudiam \dict_m(d, S)$ in
\autoref{fig:diam-mem} for a setting where $|\samplespace|=1000$, $\probap{X_A =
        1} = 0.3$ and the $\mu$-diameter of the random strategy is averaged over $100$
realisations of $S$. We first observe the drastic impact of dictionary memory on
an audit of a fixed budget: for instance, with an audit budget of 300
(representing nearly one-third of the whole input space) an optimal audit set
barely achieves a $\mu$-diameter of 1 when auditing dictionaries with memory
$m=500$. Furthermore, given a fixed audit budget, the gap between randomized and
optimal audit sets shrinks as the memory grows. This is especially striking in
low audit budget regimes, that correspond to a typical audit situation.
Moreover, for an audit budget of $100$ and memory values larger than $70\%$ the
random and optimal audit strategies have the same $\mu$-diameter. This
observation hints that \autoref{thm:shattering_implies_point_equivalence}'s
conclusions should hold for a broader set of hypothesis classes.

\subsection{Tying it all together: large capacity and auditability}\label{subsec:large_capacity_models}

We derived in \autoref{subsec:dictionaries} the exact expression of the
$\mu$-diameter for toy models able to memorize part of the input space.
Motivated by the \emph{benign overfitting} phenomenon
\cite{zhangUnderstandingDeepLearning2021,belkinReconcilingModernMachinelearning2019,arnouldInterpolationBenignRandom2023,buschjagerThereNoDoubleDescent2021},
we now consider the case of a hypothesis class that is able to perfectly fit any
subset $S \subseteq \samplespace$ of reasonable size, but require in addition
that the resulting hypothesis $h^*$ maintains good accuracy on the rest of the
dataset.

It has been observed that contrary to common knowledge on the bias-variance
tradeoff, large ML models can exhibit good generalization properties while
perfectly fitting the train data. This benign overfitting phenomenon (also
related to \emph{double descent}), is observed in models that are largely
overparametrized compared to the training data available at hand. Nevertheless,
we show in \autoref{fig:capacity} that trees and GBDTs can reach the maximum
capacity, indicating that they also can interpolate the training data. Drawing
intuition from the empirical characterization of benign overfitting in
\cite{zhangUnderstandingDeepLearning2021,belkinReconcilingModernMachinelearning2019,arnouldInterpolationBenignRandom2023,buschjagerThereNoDoubleDescent2021},
we derive the formal definition of a large capacity hypothesis class in
\autoref{def:large_capacity}.

\begin{definition}[Benign Overfitting Hypothesis class]\label{def:large_capacity}
    Consider an input space $\samplespace$, a hypothesis class $\hypotspace \subset
        \set{0,1}^\samplespace$ and a labeling $c \in \set{0,1}^\samplespace$.
    $\hypotspace$ is said to exhibit benign overfitting with respect to labeling $c$
    if there exists $d_0 \in \N_*$ and $\epsilon \in [0, 1)$ such that%
    \begin{align*}
         & \forall d \leq d_0, S \subseteq \samplespace, \sigma \in \set{0, 1}^d, \exists h \in \hypotspace,          \\
         & \begin{cases}
               \forall x_i \in S, h(x_i) = \sigma_i                      & \eqcom{(fit any train set)}     \\
               \probap{h(X) = c(X) \given X \in \snot{S}} = 1 - \epsilon & \eqcom{(with low error on $c$)}
           \end{cases}
    \end{align*}
\end{definition}

As is stands, \autoref{def:large_capacity} is tightly linked to the notion of
version space. If $\hypotspace$ exhibits overfitting, we are guaranteed that all
the version spaces $\hypotspace(h^*,S)$ (such that $\card{S} \leq d_0$) derived
from $\hypotspace$ contain a hypothesis that generalizes well on the whole
dataset. Moreover, \autoref{def:large_capacity} is the literal formalization of
the notion of benign overfitting considered in
\cite{zhangUnderstandingDeepLearning2021} and
\cite{belkinReconcilingModernMachinelearning2019}: models that can fit any
labeling --even random-- of the train set while still having a good test
performance when evaluated on the target distribution.

This definition of large capacity models enables the same analysis as in
\autoref{thm:shattering_implies_point_equivalence}, without the requirement that
the hypothesis class $\hypotspace$ spans the entire set of functions
$\set{0,1}^\samplespace$.

\begin{corollaryrep}[Benign overfitting and $\mu$-diameter]\label{corr:sensitive_attr_predictability_implies_point_equivalence}
    Let $\samplespace$ and $\hypotspace \subseteq \{0,1\}^\samplespace$ be any input
    space and hypothesis class. Assume that $\hypotspace$ exhibits benign
    overfitting with respect to the sensitive attribute $X_A$ and its opposite $1 -
        X_A$ \footnote{That is, \autoref{def:large_capacity} holds for $c = x_A$ and $c
            = 1 - x_A$}, then $\forall d \leq d_0, S \in \samplespace^d$,%
    \begin{align*}
        \mudiam \hypotspace(h^*, S) \geq & \probap{X \in S \given X_A = 1} + \probap{X \in S \given X_A = 0}     \\
                                         & - 2  \probap{X \in S} - 2 \epsilon \left(1 - \probap{X \in S} \right)
    \end{align*}
\end{corollaryrep}%
\begin{proof}
    Note $\alpha_1 = \probap{X \in \snot{S} \given X_A = 1}$ and $\alpha_0 =
        \probap{X \in \snot{S} \given X_A = 0}$. In the proof of
    \autoref{thm:shattering_implies_point_equivalence}, we established the following
    equality:
    \begin{align}
         & \mudiam \hypotspace(h^*, S)                                                                                                                                                                                    \\
         & = \colorbox{orange!20!white}{$\max_{h \in \hypotspace(h^*, S)} \left\{\alpha_1 \probap{h(X) = 1 \given X_A = 1, X \in \snot{S}}+ \alpha_0 \probap{h(X) = 0 \given X_A = 0, X \in \snot{S}}\right\}$} \nonumber \\
         & - \colorbox{blue!20!white}{$\min_{h \in \hypotspace(h^*, S)} \left\{\alpha_1 \probap{h(X) = 1 \given X_A = 1, X \in \snot{S}}+ \alpha_0 \probap{h(X) = 0 \given X_A = 0, X \in \snot{S}}\right\}$}
    \end{align}

    And
    \begin{equation}
        \colorbox{blue!20!white}{\phantom{blind}} = \alpha_1 + \alpha_0 - \max_{h \in \hypotspace(h^*, S)} \left\{\alpha_1 \probap{h(X) = 0 \given X_A = 1, X \in \snot{S}}+ \alpha_0 \probap{h(X) = 1 \given X_A = 0, X \in \snot{S}}\right\}
    \end{equation}

    Since $\hypotspace$ exhibits benign overfitting with respect to the sensitive
    attribute and $\card{S} <= d_0$, there exists $h \in \hypotspace(h^*, S)$ such
    that $\probap{h(X) = X_A \given X \in \snot{S}} = 1 - \epsilon$. Moreover,
    \begin{align}
        \probap{h(X) = X_A \given X \in \snot{S}} = & \probap{h(X) = 1 \given X \in \snot{S}, X_A = 1} \probap{X_A = 1 \given X \in \snot{S}} \nonumber                    \\
                                                    & + \probap{h(X) = 0 \given X \in \snot{S}, X_A = 0} \probap{X_A = 0 \given X \in \snot{S}}                            \\
        =                                           & \alpha_1 \frac{\probap{X_A = 1}}{\probap{X \in \snot{S}}} \probap{h(X) = 1 \given X \in \snot{S}, X_A = 1} \nonumber \\
                                                    & + \alpha_0 \frac{\probap{X_A = 0}}{\probap{X \in \snot{S}}} \probap{h(X) = 1 \given X \in \snot{S}, X_A = 0}
    \end{align}

    Since $\probap{X_A = 0} + \probap{X_A = 1} = 1$, $\probap{X_A = 0} \geq 0$ and
    $\probap{X_A = 1} \geq 0$, we have
    \begin{align}
         & \alpha_1 \probap{h(X) = 1 \given X \in \snot{S}, X_A = 1} + \alpha_ 0 \probap{h(X) = 1 \given X \in \snot{S}, X_A = 0} \nonumber                              \\
         & \geq \probap{X_A = 1} \alpha_1 \probap{h(X) = 1 \given X \in \snot{S}, X_A = 1} + \probap{X_A = 0} \alpha_ 0 \probap{h(X) = 1 \given X \in \snot{S}, X_A = 0} \\
         & = (1 - \epsilon) \probap{X \in \snot{S}}
    \end{align}

    Therefore,
    \begin{equation}
        \colorbox{orange!20!white}{\phantom{blind}} \geq (1 - \epsilon) \probap{X \in \snot{S}}
    \end{equation}

    With the same arguments, we prove
    \begin{equation}
        \colorbox{blue!20!white}{\phantom{blind}} \leq \alpha_0 + \alpha_1 - (1 - \epsilon) \probap{X \in \snot{S}}
    \end{equation}

    To conclude,
    \begin{equation}
        \mudiam \hypotspace(h^*, S) \geq 2(1 - \epsilon) \probap{X \in \snot{S}} - (\alpha_0 + \alpha_1)
    \end{equation}
\end{proof}
Observe that lower bound on the $\mu$-diameter given by
\autoref{corr:sensitive_attr_predictability_implies_point_equivalence} only
depends on the aggregated quantities $\probap{X \in \snot{S}}$, $\probap{X \in
        \snot{S} \given X_A = 1}$ and $\probap{X \in \snot{S} \given X_A = 1}$. As for
\autoref{thm:shattering_implies_point_equivalence}, this implies that no audit
method, active or not can perform better than a simple random sampling baseline
(\autoref{algo:random_audit}) with the right proportions $\beta_1$ and
$\beta_2$. The term $\probap{X \in S \given X_A = 1} + \probap{X \in S \given
        X_A = 0} - 2 \probap{X \in S}$ indicates the importance of the relative
proportion of these two audited groups in the audit set as in
\autoref{thm:shattering_implies_point_equivalence}. The term $2 \epsilon \left(1
    - \probap{X \in S} \right)$ indicates that as expected, the larger the error
rate $\epsilon$ gets, the smaller the $\mu$-diameter will be. Thus, \textbf{when
    the hypothesis class exhibits benign overfitting, all audit algorithms --active
    or not-- have at most the same manipulation-proofness guarantees as random
    sampling.} This shows that large models currently used in production are not
auditable more efficiently than by random sampling.

\section{Manipulability under random audits and model capacity}\label{sec:difficulty_capacity}

As shown in \autoref{sec:theoretical_result}, the random audit baseline is
optimal when the model has a large capacity, but has no guarantee of optimality
when the hypothesis class is constrained to lower capacities. To compare ML
algorithms in practice, we now introduce a measure of \emph{manipulability under
    random audits} and a measure of \emph{model capacity}. We will use these methods
to empirically evaluate the manipulability of auditing several models of
increasing capacities in \autoref{sec:experiments}.

\subsection{Measuring the manipulability under random audits of practical models}\label{subsec:difficulty_capacity}

The manipulability of a hypothesis class $\hypotspace$, is defined
(\autoref{eq:audit_difficulty}) as the $\mu$-diameter obtained and averaged over
audit datasets $S$ sampled by the random audit baseline
\autoref{algo:random_audit} with budget $s$.%
\begin{equation}\label{eq:audit_difficulty}
    \manipulability (\hypotspace, s) = \E[S, h^*]{\mudiam \hypotspace (S, h^*)}
\end{equation}%

\paragraph{The manipulability under random audits is a lower bound of the auditor "power"}
In a perfect situation, for any budget $s = \card{S}$, the auditor would be able
to select the audit set $S^*$ that attains the minimum $\mu$-diameter, whatever
the hypothesis class $\hypotspace$ and chosen hypothesis $h^* \in \hypotspace$
are. As explained in \autoref{subsec:computational_complexity}, this is not
possible in practice for computational reasons and thus cannot be simulated.
Thus we evaluate the manipulability under random audits with the baseline random
audit strategy (Algorithm~\ref{algo:random_audit}). Taking the expectation of
$\mudiam \hypotspace (h^*, S)$ over random audits allows to upper bound the
value of the minimum attainable $\mu$-diameter $\min_\algo \mudiam \hypotspace
    (h^*, \algo(h^*))$.

\paragraph{The manipulability under random audits is a lower bound of the platform "power"}
In a fully adversarial setting, whatever the hypothesis class $\hypotspace$, the
platform would choose the hypothesis $h^*$ that maximizes $\mudiam
    \hypotspace(h^*, S)$ for most of the audit sets $S$ the auditor could come up
with. While this would effectively be the worst case for the auditor, it is
however unlikely to happen in practice since the platform would have to balance
the maximization of the accuracy with the maximization of the $\mu$-diameter.
Therefore, we consider the more practical situation in which the platform can
freely choose the hypothesis class $\hypotspace$ but the implemented instance
$h^*$ minimizes a classical loss $L$ adapted to the model being trained (e.g.
cross-entropy or $\ell_2$ norm). This can be seen as a lower bound of the
adversarial "power" of the platform.

\subsection{Measuring the capacity of practical models}\label{subsec:model_capacity}
There are multiple operationalizations of the notion of capacity, from
theoretically-rooted metrics such as the VC dimension
\cite{vapnikUniformConvergenceRelative1971} or Rademacher complexity
\cite{shalev-shwartzUnderstandingMachineLearning2014}, to more empirical
definition such as the number of iterations until overfitting
\cite{zhangUnderstandingDeepLearning2021}. The interplay between VC-dimension
and manipulability under random audits is already pointed out in
\cite{yanActiveFairnessAuditing2022}, where it is observed that models of
VC-dimension higher than $1,600$ have a high manipulability under random audits.

Unfortunately, the VC dimension of a class is difficult to estimate in practical
settings. Instead, the empirical Rademacher complexity
(\autoref{eq:empirical_rademacher}) is leveraged to quantify the capacity of the
studied hypothesis classes. Informally in our setting, a hypothesis class has a
high Rademacher complexity if whatever the labels and size of an audit set $S$,
there exists an instance $h_S \in \hypotspace$ that fits those labels on $S$
with high accuracy. To avoid threshold effects in our experiments, we average
the complexity over different sizes of $D$ considered in the Rademacher metric
(\autoref{eq:capacity}). Formally:%
\begin{align}
    R_m (\hypotspace \circ D) & = \frac{1}{m} \E[\bm{\sigma} \sim \{\pm 1\}^m]{
        \sup_{h \in \hypotspace} \sum_{x_i \in D} \sigma_i h(x_i)
    }\label{eq:empirical_rademacher}                                            \\ \capacity(\hypotspace) & = \E[\substack{D
    \sim \samplespace^m \;\;\;                                                  \\ m \sim \llbracket
    \card{\samplespace}\rrbracket}]{R_m(\hypotspace \circ D)} \label{eq:capacity}
\end{align}%

\begin{figure}
    \centering
    \includegraphics{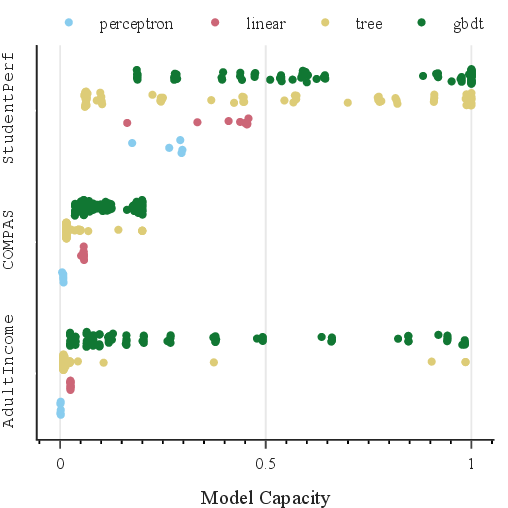}
    \caption{%
        Distribution of the capacity (horizontal axis) for different hyperparameters
        choices on the three datasets (vertical axis). Each model is trained with
        different hyperparameter values with each couple (model, hyperparameter)
        representing a different hypothesis class $\hypotspace$. For each (model,
        hyperparameter) couple, the empirical Rademacher values $R_m (\hypotspace \circ
            D)$ are averaged over $15$ realizations of $D$ and $\sigma_i$ before computing
        the model capacity.
    }
    \label{fig:capacity}
\end{figure}

\begin{figure}
    \centering
    \includegraphics{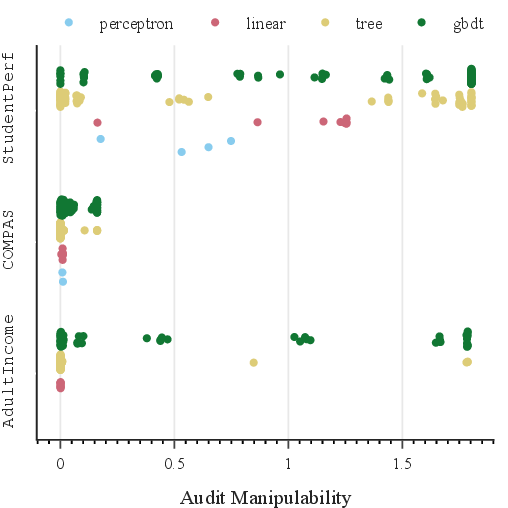}
    \caption{%
        Distribution of the $\manipulability$ (manipulability under random audits)
        values (horizontal axis) of different models $\family$ on a selection of
        datasets (vertical axis). Each bar represents a different model $\family$
        (trees, linear models, ...). Each model is trained with different hyperparameter
        values with each couple (model, hyperparameter) representing a different
        hypothesis class $\hypotspace$. For each dataset, the size of the audit set is
        set to $10\%$ of the dataset size: $\card{S} = 0.1 \card{\samplespace}$. For
        each (model, hyperparameter) couple, the $\mu$-diameter are averaged over $15$
        audit datasets before computing the manipulability.
    } \label{fig:mu-diam}
\end{figure}

\section{Experiments}\label{sec:experiments}
In this section, we explore the relation of the manipulability under random
audits (\autoref{eq:audit_difficulty}) with the capacity of hypothesis classes
(\autoref{eq:capacity}). The following experiments were run on three tabular
datasets: \student~\cite{cortezUsingDataMining2008},
\compas~\cite{larsonHowWeAnalyzed2016} and
\adult~\cite{dingRetiringAdultNew2021}. Dataset statistics and considered tasks
are presented in \autoref{tab:dataset_stats}. Neural methods on tabular data are
still outperformed by tree methods \cite{grinsztajnWhyTreebasedModels2022}. We
thus choose to focus our study on the four following models: linear models,
perceptrons, decision trees and gradient-boosted trees. Similar to
\cite{grinsztajnWhyTreebasedModels2022}, we selected a range of hyperparameters
for each model and sampled a total of $500$ hyperparameters over the $4$ models.
In previous sections, we stated results with respect to a given hypothesis class
$\hypotspace$. In the following experiments, a hypothesis class $\hypotspace$
represents a couple (model, hyperparameters). Thus, a model represents a family
of hypothesis classes $\family = (\hypotspace_1, \dots, \hypotspace_f)$, each
hypothesis class $\hypotspace_i$ being associated with a hyperparameters tuple.

The hyperparameters and their value range are presented in
\autoref{tab:models_hparams}. For each model, we created a grid with all the
possible combinations of hyperparameter values and ran our experiments on all
the resulting (model, hyperparameter) couples. The code needed to run the
experiment, the hyperparameters, the data we obtained and the code to reproduce
the figures will be made available upon publication.

\begin{table}
    \caption{Datasets stats}\label{tab:dataset_stats}
    \centering
    \begin{tabular}{r|>{\raggedleft}p{.6cm} >{\raggedleft}p{.6cm} p{4cm}}
        dataset  & Size $n$ & Features $d$ & Task                                 \\ \hline \\[-5pt]
        \student & 395      & 43           & Predict if students pass the exam    \\
        \compas  & 6172     & 21           & Predict subject recidivism           \\
        \adult   & $22,268$ & 10           & Predict if income is $\geq \$50,000$
    \end{tabular}
\end{table}

\begin{table}
    \centering
    \caption[Value range for the hyperparameters of the models used in the experiments.]{Value range for the hyperparameters of the models used in the experiments.}
    \label{tab:models_hparams}
    \begin{tabular}{>{\raggedleft}p{2.5cm}|p{5.5cm}}
        Model \& hyperparameters                 & Value range                                                           \\

        \hline \phantom{blind}                                                                                           \\[-5pt]
        \raggedright\textsc{\textbf{linear}}     &                                                                       \\
        penalty                                  & \texttt{(None, l2)}                                                   \\
        C                                        & \texttt{(0.001, 0.01, 0.1, 1, 10, 100, 1000, 10000)}                  \\

        \raggedright\textsc{\textbf{perceptron}} &                                                                       \\
        penalty                                  & \texttt{(l2, )}                                                       \\
        alpha                                    & \texttt{(1e-06, 1e-05, 0.0001, 0.001, 0.01)}                          \\

        \raggedright\textsc{\textbf{tree}}       &                                                                       \\
        max\_depth                               & \texttt{(2, 4, 8, 16, 32, 64, 128)}                                   \\
        ccp\_alpha                               & \texttt{(0.001, 0.003, 0.005, 0.007, 0.01, 0.05, 0.1, 0.2, 0.5, 0.0)} \\

        \raggedright\textsc{\textbf{gbdt}}       &                                                                       \\
        max\_depth                               & \texttt{(1, 2, 4, 8)}                                                 \\
        n\_estimators                            & \texttt{(100, 200, 500)}                                              \\
        reg\_lambda                              & \texttt{(0.0, 1e-6, 1e-3, 0.1, 1.0, 1e6, 1e7)}                        \\
        max\_leaves                              & \texttt{(0,)}                                                         \\
        learning\_rate                           & \texttt{(0.3,)}                                                       \\
        gamma                                    & \texttt{(0.0,)}                                                       \\
        min\_child\_weight                       & \texttt{(0.0,)}                                                       \\
        max\_delta\_step                         & \texttt{(0.0,)}                                                       \\
        subsample                                & \texttt{(1.0,)}                                                       \\
        reg\_alpha                               & \texttt{(0.0,)}                                                       \\
        early\_stopping\_rounds                  & \texttt{(None,)}
    \end{tabular}
\end{table}

\subsection{Simulating hypothesis spaces with a broad range of manipulability and capacity}\label{subsec:difficulty_capacity_range}

In \autoref{fig:mu-diam}, we plot the manipulability under random audits of
different hypothesis classes. These classes are constructed by using multiple
hyperparameters for each family $\family$ listed in
\autoref{tab:models_hparams}; each dot then represents a specific (family,
hyperparameter set) couple. On one hand, for large datasets (such as \adult and
\compas), we observe that simpler models (linear, perceptron) have a very low
manipulability, no matter the hyperparameter set used. On the other hand, for
smaller datasets (such as \student), smaller models (such as linear models or
perceptrons) can also fit the data hence also becoming harder to audit.

Similarly, in \autoref{fig:capacity}, we plot the capacity of the simulated
hypothesis classes on \adult, \compas~and \student. As discussed before, it can
be observed that for \adult~and \student~datasets, tree-based models reach the
maximum capacity value of $1$. However, on the \compas~dataset all hypothesis
classes exhibit capacity values that do not exceed $0.2$ points. This has been
observed before \cite{dresselAccuracyFairnessLimits2018} and does not affect our
main argument on the link between model capacity and manipulability.

\subsection{Model capacity conditions manipulability}\label{subsec:mu-diam_vs_capacity}
\begin{figure*}
    \centering
    \includegraphics{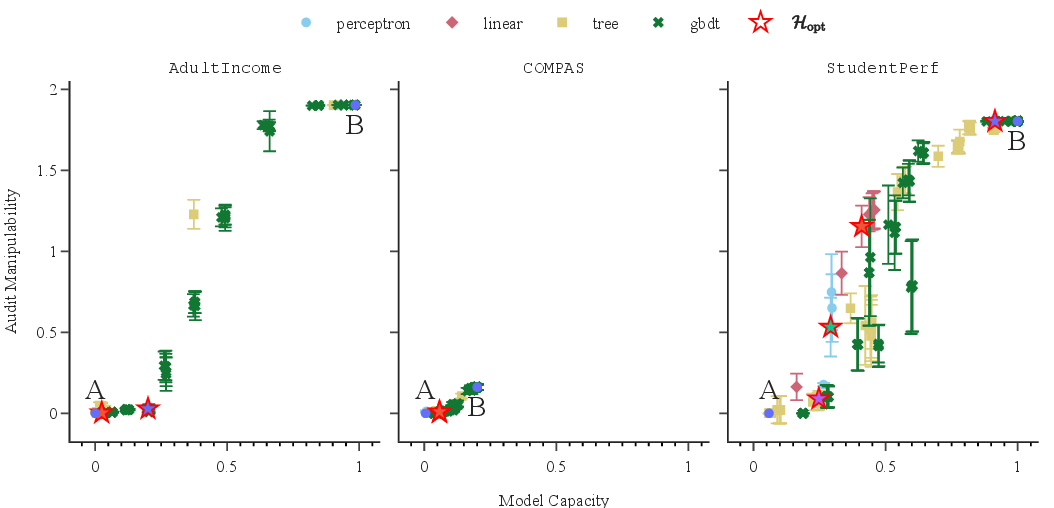}
    \caption{
        Distribution of the manipulability under random audits values (vertical axis) of
        different models versus their capacity (horizontal axis) on a selection of
        datasets. Each point represents a couple (model, hyperparameter). For each
        dataset, the size of the audit set is set to $10\%$ of the dataset size:
        $\card{S} = 0.1 \card{\samplespace}$. For each (model, hyperparameter) couple,
        the \manipulability is averaged over $15$ audit datasets, and the capacity is
        computed over $30$ randomizations of the dataset labels. The error bars
        represent the standard deviation. }

    \label{fig:mu-diam_vs_capacity}
\end{figure*}

In \autoref{subsec:difficulty_capacity} we compared different models and how
difficult they were to audit, depending on the chosen hyperparameters. We now
take a closer look at the impact of a model's capacity on its manipulability
under random audits, in an attempt to confirm the link between both concepts. We
plot in \autoref{fig:mu-diam_vs_capacity} the relation between the capacity of a
hypothesis class and its manipulability under random audits. Points also
represent (model, hyperparameter) couples, while the vertical error bars
represent the standard deviation of the $\mu$-diameter values for different
random audit sets $S$.

Consistent with the intuition and results developed until now, we observe that
for all the datasets, the manipulability under random audits increases with the
capacity of the hypothesis class. While on both \adult~and \student, the
$\mu$-diameter reaches the maximum capacity value at almost $2$, for \compas,
the effect is not as dramatic. To highlight the connection between the results
exposed in \autoref{sec:theoretical_result} and the empirical relation found
between model capacity and manipulability under random audits, we focus next on
two specific points, marked with the letters $A$ and $B$ in
\autoref{fig:mu-diam_vs_capacity}.

First, consider the point $A = (\capacity \simeq 0, \manipulability \simeq 0)$.
For a hypothesis class to have a null capacity, it has to have null Rademacher
complexity on any subset of the sample space. This is verified by models that
perform no better than random labels generation. Since the value of $\mu(h,
    \samplespace)$ of any instance of such hypothesis class is only determined by
the ratio of samples with a positive sensitive attribute, the $\mu$-diameter of
such hypothesis class is null. This is why in \autoref{fig:mu-diam_vs_capacity},
models with near-zero capacity have a very low (if not null) manipulability
under random audits.

The second notable point is $B = (\capacity = \capacity_{\max}, \manipulability
    = \manipulability_{\max})$. Any hypothesis with a unitary capacity has a unitary
Rademacher complexity for any dataset size $s$ and thus shatters any subset of
$\samplespace$. Therefore, at point B,
\autoref{thm:shattering_implies_point_equivalence}'s hypothesis $\hypotspace =
    \set{0, 1}^\samplespace$ holds. This means that hypothesis classes that are
characterized by this point cannot be audited more efficiently than by a random
audit strategy. It follows that (at least on \student~and \adult) the platform
can always choose a hypothesis class that cannot be audited efficiently by any
strategy, forcing the auditor to prompt most of the input space to obtain
robustness guarantees.

\paragraph*{Generalization versus diameter} We saw that by choosing the right
hypothesis class (that is, the right set of hyperparameters), the platform can
easily evade the audit. However, in practice the choice of hypothesis class is
also guided by a classical train-dev-test separation, choosing the
hyperparameter set that generalizes best. What is the typical $\mu$-diameter of
hypotheses classes that generalize well? To answer this question, we simulate a
5-fold hyperparameter optimization procedure. For each family of models, we
denote $\hypotspace_\text{opt}$ the hypothesis class with the set of
hyperparameters that minimize the 5-fold average test loss in its model family
$\family$. For each model family, $\hypotspace_\text{opt}$ is differentiated in
\autoref{fig:mu-diam_vs_capacity} by a star marker with red edges.
Interestingly, for \compas~and \adult~datasets and for all model families, the
generalization-optimal hypothesis classes $\hypotspace_\text{opt}$ have a
relatively low capacity compared to the maximum achievable capacity, especially
for tree-based models. For the \student~dataset, the results are more nuanced,
most likely because the dataset has a limited size, which implies that it is
simpler to reach high capacity values.%

As a glimmer of hope, from point $A$ to $B$, there is a range of hypothesis
classes for which the random strategy could be improved as seen by the size of
the $y$-axis error bars. Overall, the hypothesis classes that are most likely to
be implemented by faithful platforms (the hypothesis classes that generalize
well) are already straightforward to audit (they have a $\manipulability \approx
    0$). Yet, unfaithful platforms wanting to game the audit can always choose a
hypothesis class that forces the auditor to issue a lot of queries to reach
higher manipulation-proofness guarantees.

\subsection{The cost of exhausting the auditor}\label{subsec:cost_of_exhaustion}

\begin{figure}
    \centering
    \includegraphics{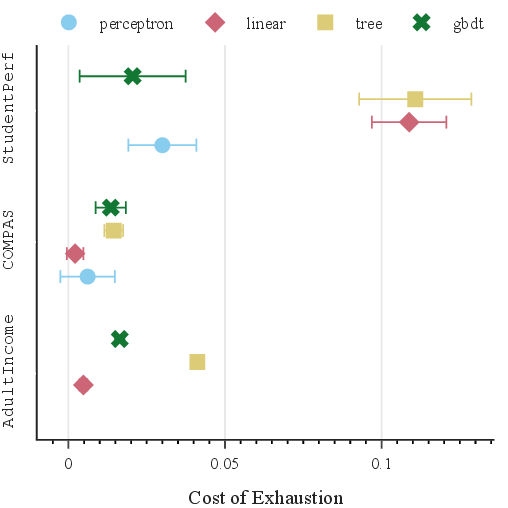}
    \caption{%
        Distribution of the \textit{cost of exhaustion} for the four model families
        (perceptron, linear, tree and GBDT) on the three considered datasets. The error
        bars show the $95\%$ confidence interval on the values of the difference of
        $\text{Accuracy}_\text{test}$ for the best hypotheses in
        $\hypotspace^\text{acc}$ and $\hypotspace^\mu$. For all models, on all datasets
        (except for trees and linear models on \student), the cost of exhaustion is
        below $1\%$. Trees are the models with the highest cost of exhaustion, while for
        all the other models, the cost of exhaustion remains relatively low (in
        particular for the large capacities GDBTs), indicating a negligible accuracy
        cost for audit evasion.
    }
    \label{fig:cost_of_exhaustion}
\end{figure}

We observed in \autoref{sec:theoretical_result} and
\autoref{subsec:difficulty_capacity} that the hypothesis classes that are the
hardest to audit are those with the largest capacity. Yet, we also observed that
the hypothesis classes most likely to be implemented (i.e. the ones with the
highest generalization) have a low $\mu$-diameter and are not those with high
capacity. In the manipulation-proof framework of
\cite{yanActiveFairnessAuditing2022} we operate in, the platform chooses the
hypothesis class without constraints before disclosing it to the auditor.
Therefore, when choosing a specific model family $\family$, a malicious platform
would have the possibility to trade performance (i.e. generalization capability)
with the ability to attempt audit evasion. To understand the trade-offs involved
in balancing these two objectives, we introduce the notion of
$\costexh(\family)$ of a model family $\family$.

A model family $\family = \set{\hypotspace_1, \dots, \hypotspace_F}$ is a set of
hypothesis classes. The family $\family$ of decision trees where each hypothesis
class $\hypotspace_i$ corresponds to a maximum depth value $i$ is an example of
model family. To define the $\text{CostOfExhaustion}$ metric, we first introduce
two particular hypothesis ($\hypotspace^\text{acc}$ and $\hypotspace^\mu$)
classes of $\family$. $\hypotspace^\text{acc}$ is the hypothesis class in
$\family$ with the best trained test accuracy:%
\begin{equation}
    \hypotspace^\text{acc} = \argmax_{\hypotspace \in \family} \max_{h \in
        \hypotspace} \text{Accuracy}_\text{test}(h, \samplespace)
\end{equation}
Assuming that an honest platform chooses its hypothesis class based on
generalization capabilities, $\hypotspace^\text{acc}$ is the hypothesis class an
honest platform would actually choose. Then, define the hypothesis class in
$\family$ with the largest manipulability (for a fixed audit budget $s$):%
\begin{equation}
    \hypotspace^\mu =
    \argmax_{\hypotspace \in \family} \manipulability(\hypotspace, s)
\end{equation}
Should a platform try to escape audits at a low cost, they would try to find a
hypothesis class whose optimal hypothesis $h^*$ leads to a high $\mu$-diameter.
Thus, the cost of exhaustion is the accuracy cost of using the hypothesis class
$\hypotspace^\mu$ compared to using $\hypotspace^\text{acc}$:
\begin{align}
     & \text{CostOfExhaustion}(\family) = \nonumber                                                                                                                     \\
     & \quad \max_{h \in \mathcal{H}^\text{acc}} \text{Accuracy}_\text{test}(h, \mathcal{X}) - \max_{h \in \mathcal{H}^\mu} \text{Accuracy}_\text{test}(h, \mathcal{X})
\end{align}

The cost of exhaustion is plotted in \autoref{fig:cost_of_exhaustion}, for the
four model families already considered, on the three datasets. The error bars
show the $95\%$ confidence interval on the values of the difference of
$\text{Accuracy}_\text{test}$ for the best hypotheses in
$\hypotspace^\text{acc}$ and $\hypotspace^\mu$. For all models, on all
considered datasets (except for trees and linear models on the dataset
\student), the cost of exhaustion is below $1\%$. Trees are the models with the
highest cost of exhaustion. In fact, as we observed in
\autoref{fig:mu-diam_vs_capacity}, given enough capacity, trees can reach the
maximum manipulability under random audits. Yet, it is known that without
regularization, complex trees can easily overfit the training data, thus
lowering the max test accuracy of the $\hypotspace^\mu$ class compared to the
max test accuracy of $\hypotspace^\text{acc}$. On the other hand, the models
with the lowest cost of exhaustion (except on \student) are linear models. As
observed on \autoref{fig:capacity}, for all datasets, linear models span a small
portion of the capacity range (around $.1$ points for \student~and less than
$.01$ points for \compas~and \adult), compared to larger models (e.g. GBDTs)
which cover almost the entire capacity range on \student~and \adult. This result
is challenging for the existence of efficient audits in the manipulation-proof
framework. In fact, the witnessed low cost of exhaustion for larger capacity
models indicates that platforms may evade audits at the cost of a minor loss in
accuracy.

\subsection{Effects of the audit set size}\label{subsec:audit_size}
\begin{figure*}
    \centering
    \includegraphics[width=\textwidth]{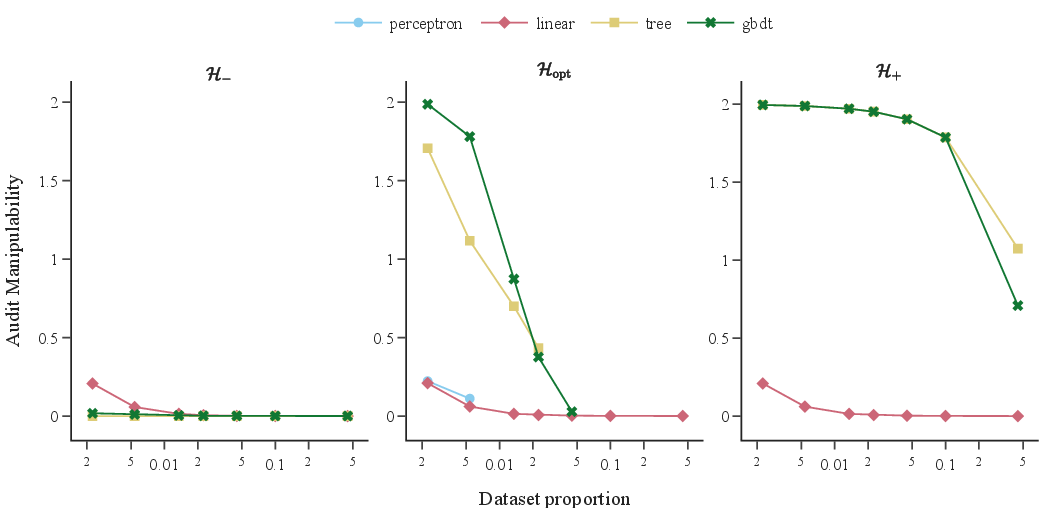}
    \caption{
        Evolution of the $\mu$-diameter with the size of the audit set $S$ represented
        as a proportion of the total dataset size for the \adult~dataset. Each line
        represents an audited model, whose hyperparameters are either tuned for the best
        generalization, either tuned for the highest capacity or tuned for the lowest
        capacity. For each (model, hyperparameter) couple, the $\mu$-diameter is
        averaged over $15$ audit datasets. }

    \label{fig:mu-diam_vs_proportion_adult}
\end{figure*}

In this section, we experiment with different sizes of audit dataset and show
that our conclusions do not change with the change in dataset size (we had
$\card{S} = .1 \card{\samplespace}$ in previous experiments). To do so, we
select three different hypotheses classes for each model family. We choose the
hypothesis class that generalizes best $\hypotspace_\text{opt}$, the hypothesis
class with the lowest capacity $\hypotspace_-$ and with the highest capacity
$\hypotspace_+$. In \autoref{fig:mu-diam_vs_proportion_adult} we show the audit
difficulty of each hypothesis class against the size of the audit dataset
$\card{S}$. The results indicate that there is no significant inversion of the
manipulability under random audits between the various hypotheses in the range
of interest. Results in \autoref{fig:mu-diam_vs_proportion_adult} are shown only
for the \adult~dataset. The results for the other datasets are showed in the
Appendix, in Figures \ref{fig:mu-diam_vs_proportion_compas} and
\ref{fig:mu-diam_vs_proportion_student_perf}, which to the same conclusion.

\section{Related work}\label{sec:related_work}

The problem of manipulation-proof auditing and more generally black-box, remote,
and robust property verification of ML platforms arises from the need to enforce
regulations. As an example, consider the European Union. Classical fairness
regulation of online ML models mainly comes from the \emph{Racial Equality
    Directive}~\cite{CouncilDirective20002000}, the \emph{Framework Equality
    Directive}~\cite{CouncilDirective20002000a} and the \emph{Gender Equality
    Directives}~\cite{CouncilDirective20042004, Directive2006542006}. Recently, the
EU set out to create regulations specific to online platforms. These are the
\emph{AI Act}~\cite{ProposalREGULATIONEUROPEAN2021}, the \emph{Digital Services
    Act}~\cite{RegulationEU20222022} and the \emph{Digital Markets
    Act}~\cite{RegulationEU20222022a}. These directives provide a legal framework
that prescribes what online platforms may and may not do, but offer little to
verify that these rules are respected in practice. The manipulation-proof
framework is a first attempt to provide operational solutions that can detect
when platforms do not follow the law.

In addition, our results are mostly related to the following lines of work.

\paragraph{Algorithm auditing}
The field of algorithm auditing is interested in understanding the impact of
algorithms on the lives or the people impacted by those algorithms' decisions.
In practice, auditing algorithms \emph{in vivo} (that is as they are deployed in
online services) is challenging because they constantly evolve, mostly without
records \cite{metaxaAuditingAlgorithmsUnderstanding2021}. For a survey on
examples of published academic audits of decision systems, refer to
\cite{bandyProblematicMachineBehavior2021}. Moreover, because it is impossible
for researchers or regulators to audit each automated decision system, it has
been observed that most of the recent discoveries of problematic algorithm
behavior have surfaced thanks to users of those
systems~\cite{devosUserDrivenAlgorithmAuditing2022,
    dengUnderstandingPracticesChallenges2023}. Again, after a problematic algorithm
behavior has been detected and after a court decision has been made, we still
need to be able to monitor that this decision is respected.

\paragraph{Audit metrics and audit design}
With the advent of broadly publicized algorithm audits such as COMPAS
\cite{larsonHowWeAnalyzed2016} or Reuters' study on Amazon's recruiting tool
\cite{dastinAmazonScrapsSecret2018}, there has been an effort to devise metrics
and their interpretations to better understand the impact of algorithms on their
users. Most of the effort has been directed towards the operationalization of
fairness values into the ML framework~\cite{barocasFairnessMachineLearning2023}.
Classical fairness measures include Demographic Parity
\cite{caldersBuildingClassifiersIndependency2009}, Equalized Odds
\cite{hardtEqualityOpportunitySupervised2016}, Equal
Opportunity~\cite{hardtEqualityOpportunitySupervised2016} or Predictive Parity
\cite{corbett-daviesAlgorithmicDecisionMaking2017}. All of these measures
encompass different visions of fairness and choosing one versus the other has
political implications on the considered notion of fairness
\cite{heidariMoralFrameworkUnderstanding2019,
    arvindnarayananTutorial21Fairness2018}. While still marginal, some works are
interested in other aspects of the audit of AI algorithms. For example,
\cite{rastegarpanahAuditingBlackBoxPrediction2021} is interested in the
verification that online platforms comply with the Data Minimization Principle.
Another interesting work~\cite{luGeneralFrameworkAuditing2022} considers the
problem of automatically auditing the privacy guarantees offered by AI
algorithms. However, most of the presented works do not yet consider the
possibility of the platform gaming their audit.

\paragraph{Robust verification}\label{subsec:related_robust_audit}
The literature on robust auditing is still in its infancy. The
manipulation-proof \cite{yanActiveFairnessAuditing2022} framework has only
recently been introduced. However, with its goal of efficiently choosing the
next audit query based on previous queries and the associated outputs of the
API, the manipulation-proof framework exhibits clear links with the active
learning literature \cite{hannekeTeachingDimensionComplexity2007,
    dasguptaTeachingBlackboxLearner2019}. With the aim of finding methods to ease
the audit, \cite{yadavXAuditTheoreticalLook2023} showed that the explanation
provided by the platform can greatly improve the robustness of audits. For
example, they show that for linear classifiers, a single result along its
counterfactual explanation allows to totally characterize the model. Our work
does not assume that the auditor has access to explanations. It is likely that
faithful explanation could lead to audit algorithms with increased MP
guarantees. On another line of works,
\cite{shamsabadiConfidentialPROFITTConfidentialPROof2023} and
\cite{goldwasserInteractiveProofsVerifying2021} suggest instantiating an audit
protocol in which both the platform and the auditor would be active, drawing
inspiration from zero-knowledge proofs and interactive verification protocols.

\paragraph{Benign overfitting and model capacity}\label{par:related_work_capacity}
As we proved in this work, manipulability under random audits has deep
connections with model capacity and their ability to perfectly fit arbitrary
datasets. Classical metrics that capture the notion of model capacity include
the VC-dimension \cite{vapnikUniformConvergenceRelative1971} or the Rademacher
Complexity (which we used for its usability in practice)
\cite{shalev-shwartzUnderstandingMachineLearning2014}. Moreover, our experiments
on the link between auditability and model capacity have been motivated by the
recent finding that larger models can fit the training dataset perfectly while
still showing good generalization properties
\cite{zhangUnderstandingDeepLearning2021}. This effect has been observed for
linear models \cite{bartlettBenignOverfittingLinear2020}, Support Vector
Machines \cite{wangBenignOverfittingMulticlass2021a} and Decision Trees
\cite{arnouldInterpolationBenignRandom2023}. In the manipulation-proof audit
setting, we show that this type of behavior is very problematic. In fact, if a
model is able to fit any audit set and yet keep its generalization performance,
platforms do not even have to lie to the auditor. They just have to train their
model to give the answers the auditor expects on their audit set. Then, the
platform can define any objective for the rest of the input space, even if it
does not align with the auditor's metric.

Interestingly, the connection between model capacity and audit query complexity
is not limited to manipulation-proof estimation of parity measures. In their
work on certified feature sensitivity auditing
\cite{yadavXAuditTheoreticalLook2023},
\citeauthor{yadavXAuditTheoreticalLook2023} provide an algorithm to audit
feature sensitivity for decision trees whose query complexity grows linearly
with the capacity (number of nodes) of the tree.

\section{Conclusion and discussions}\label{sec:conclusion}
The introduction of the \emph{manipulation-proofness} framework
\cite{yanActiveFairnessAuditing2022} has certainly been an important step for
auditors to start understanding that algorithmic audits can suffer from platform
manipulations and what cost that brings along.

In this work, we conducted a thorough exploration of the concept of
manipulation-proofness. We derived theoretical conditions on the hypothesis
class implemented by the platform for the impossibility of efficient
manipulation-proof audits. We carried out a thorough experimental validation on
the \emph{manipulability under random audits} of state-of-the-art models for
tabular data. Our results draw a connection between the capacity of the audited
model and the difficulty of the audit task.

We now discuss some countermeasures to improve the audit robustness. A promising
line of work is to require platforms to provide certificates. Since the goal of
certificates is to provide a cheap verification procedure (at the cost of a
potentially high certificate generation cost), this would shift the
computational burden to the platform. One example of a fairness certificate was
provided in \cite{shamsabadiConfidentialPROFITTConfidentialPROof2023}. Such
extended assumptions (over mere black box audits) are certainly an interesting
research line for future works.

In the end, when implementing large-capacity models, a platform can always game
the audit without sacrificing too much accuracy.
\llabel{l:conclusion_regulator_access} We believe that this demonstrates the
limitations of black-box auditing for regulation, even when the hypothesis class
used by the platform is known to the regulator. We claim that regulators should
be given more than black-box access to AI models as part of the audit procedure
or that they should explore certification-based audits such as
\cite{shamsabadiConfidentialPROFITTConfidentialPROof2023}. Therefore, we urge
the community to participate in the search for audit frameworks that are both
exploitable in practice and also supported by theoretical guarantees.

\printbibliography

\newpage
\appendix

\section{Effect of the audit dataset size}
\begin{figure*}
    \centering
    \includegraphics[width=\textwidth]{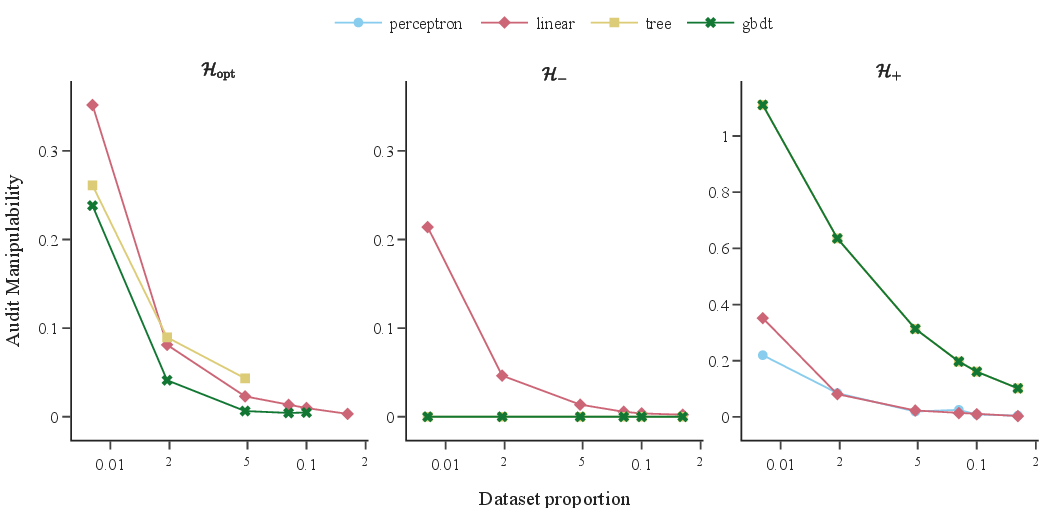}
    \caption{
        Evolution of the $\mu$-diameter with the size of the audit set $S$ represented
        as a proportion of the total dataset size for the \adult~dataset. Each line
        represents an audited model, whose hyperparameters are either tuned for the best
        generalization, either tuned for the highest capacity or tuned for the lowest
        capacity. For each (model, hyperparameter) couple, the $\mu$-diameter is
        averaged over $15$ audit datasets. }

    \label{fig:mu-diam_vs_proportion_compas}
\end{figure*}

\begin{figure*}
    \centering
    \includegraphics[width=\textwidth]{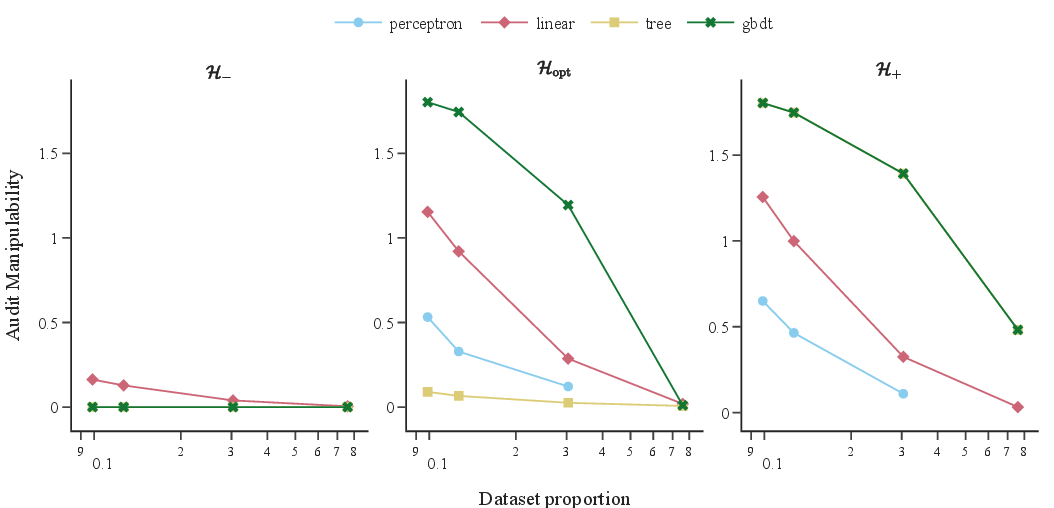}
    \caption{
        Evolution of the $\mu$-diameter with the size of the audit set $S$ represented
        as a proportion of the total dataset size for the \adult~dataset. Each line
        represents an audited model, whose hyperparameters are either tuned for the best
        generalization, either tuned for the highest capacity or tuned for the lowest
        capacity. For each (model, hyperparameter) couple, the $\mu$-diameter is
        averaged over $15$ audit datasets. }

    \label{fig:mu-diam_vs_proportion_student_perf}
\end{figure*}

\begin{toappendix}
    \section{How is the $\mu$-diameter measured in practice}\label{sec:mu_diam_computation}

    As originally defined in \cite{yanActiveFairnessAuditing2022} and following the
    definition of the $\mu$-diameter, the evaluation of $\text{diam}_\mu(S, h^*)$
    requires to solve the following optimization problem:
    \begin{align}
        \max_{h, h^\prime} \quad & |\mu(h, S) - \mu(h^\prime)|                       \\
        \text{subject to} \quad  & h(x) = h^\prime(x) = h^*(x) \quad \forall x \in S
    \end{align}

    This problem be separated in two optimization problems: the
    maximization/minimization over $h \in \hypotspace$ of $\mu(h, S)$ under the
    constraint that $\forall x \in S, h(x) = h^*(x)$.
    \begin{align}
        \max_{h} / \min_{h} \quad & \mu(h, S)                           \\
        \text{subject to} \quad   & h(x) = h^*(x) \quad \forall x \in S
    \end{align}

    As proposed by \cite{yanActiveFairnessAuditing2022}, we use the method
    introduced by \cite{agarwalReductionsApproachFair2018} to reframe this
    constrained optimization problem as a sequence of weighted classification tasks.
    Then, we use off-the-self estimators from scikit-learn and XGBoost to perform
    the optimization with the appropriate weights.
\end{toappendix}

\end{document}